\documentclass[10pt,twocolumn,letterpaper]{article}

\usepackage{cvpr}              % To produce the CAMERA-READY version
\usepackage{times}
\usepackage{epsfig}
\usepackage{graphicx}
\usepackage{amsmath}
\usepackage{amssymb}
\usepackage{booktabs}
\usepackage{float}
\usepackage{nicefrac}
\usepackage{boldline}
\usepackage{multirow}
\usepackage{color}
\usepackage{arydshln}

\usepackage[pagebackref,breaklinks,colorlinks]{hyperref}

\usepackage[capitalize]{cleveref}
\crefname{section}{Sec.}{Secs.}
\Crefname{section}{Section}{Sections}
\Crefname{table}{Table}{Tables}
\crefname{table}{Tab.}{Tabs.}

\def\cvprPaperID{3539} % *** Enter the CVPR Paper ID here
\def\confName{CVPR}
\def\confYear{2022}

\begin{document}

\title{Bridging Global Context Interactions for High-Fidelity Image Completion}

\author{Chuanxia Zheng \qquad Tat-Jen Cham\\
School of Computer Science and Engineering\\
Nanyang Technological University, Singapore \\
{\tt\small \{chuanxia001,astjcham\}@ntu.edu.sg}
\and
Jianfei Cai \qquad Dinh Phung \\
Department of Data Science \& AI\\
Monash University, Australia\\
{\tt\small  \{Jianfei.Cai,dinh.phung\}@monash.edu}
}

\twocolumn[{%
	\maketitle
	\vspace*{-1.0cm}
	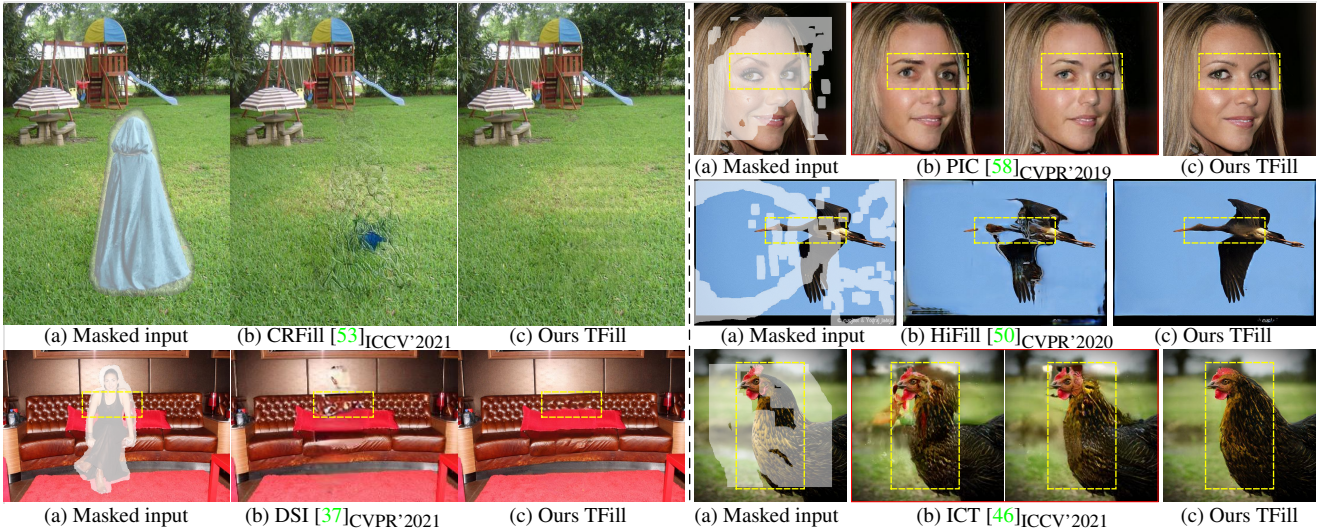
\begin{figure}[H]
		\centering
		\hsize=\textwidth % cvpr 需要
		\includegraphics[width=\textwidth]{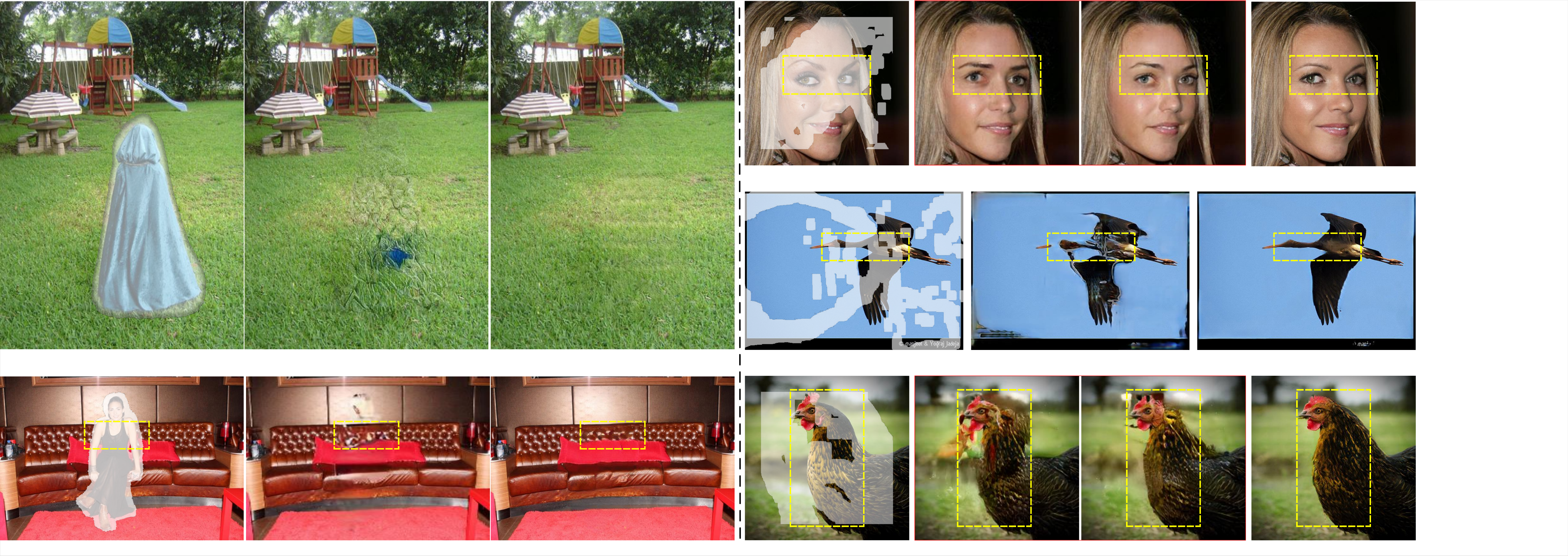}
		 \begin{picture}(0,0)
            \put(-232,73){\footnotesize (a) Masked input}
            \put(-159,73){\footnotesize  (b) CRFill~\cite{zeng2021generative}$_{\text{\scriptsize{ICCV'2021}}}$}
            \put(-56,73){\footnotesize  (c) Ours TFill}
            \put(14,137){\footnotesize (a) Masked input}
            \put(98,137){\footnotesize  (b) PIC~\cite{Zheng_2019_CVPR}$_{\text{\scriptsize{CVPR'2019}}}$}
            \put(198,137){\footnotesize  (c) Ours TFill}
            \put(24,73){\footnotesize (a) Masked input}
            \put(92,73){\footnotesize  (b) HiFill~\cite{yi2020contextual}$_{\text{\scriptsize{CVPR'2020}}}$}
            \put(188,73){\footnotesize  (c) Ours TFill}
            \put(-232,5){\footnotesize (a) Masked input}
            \put(-156,5){\footnotesize  (b)
            DSI~\cite{peng2021generating}$_{\text{\scriptsize{CVPR'2021}}}$}
            \put(-56,5){\footnotesize  (c) Ours TFill}
            \put(14,5){\footnotesize (a) Masked input}
            \put(98,5){\footnotesize  (b) ICT~\cite{Wan_2021_ICCV}$_{\text{\scriptsize{ICCV'2021}}}$}
            \put(198,5){\footnotesize  (c) Ours TFill}
        \end{picture}
    \vspace{-0.4cm}
		\caption{\textbf{Example completion results of our method on different sceneries with various masks} (missing regions shown in white, a transparency ratio is set for better visualization). Our TFill model not only effectively removes large objects (left), but also infers reasonable contents and plausible appearances for semantical image completion on various settings (right). (\textcolor[RGB]{255,0,0}{Zoom in to see the details.})}
		\label{fig:example}
	\end{figure}
}]

\maketitle

%%%%%%%%% ABSTRACT
\begin{abstract}
Bridging global context interactions correctly is important for high-fidelity image completion with large masks. Previous methods attempting this via deep or large receptive field (RF) convolutions cannot escape from the dominance of nearby interactions, which may be inferior. In this paper, we propose to treat image completion as a directionless sequence-to-sequence prediction task, and deploy a transformer to directly capture long-range dependence in the encoder. Crucially, we employ a restrictive CNN with small and non-overlapping RF for weighted token representation, which allows the transformer to explicitly model the long-range visible context relations with equal importance in all layers, without implicitly confounding neighboring tokens when larger RFs are used. To improve appearance consistency between visible and generated regions, a novel attention-aware layer (AAL) is introduced to better exploit distantly related high-frequency features. Overall, extensive experiments demonstrate superior performance compared to state-of-the-art methods on several datasets. Code is available at \href{https://github.com/lyndonzheng/TFill}{https://github.com/lyndonzheng/TFill}. 
\end{abstract}

%%%%%%%%% BODY TEXT
\section{Introduction}
\label{sec:intro}

Image completion refers to the task of filling reasonable content with photorealistic appearance into missing regions, conditioned on partially visible information (as shown in Fig.\ \ref{fig:example}). Earlier methods infer the pixels of missing regions by propagating pieces from neighboring visible regions \cite{bertalmio2000image,ballester2001filling,criminisi2004region,barnes2009patchmatch}, while more recent ones directly learn to generate content and appearance using deep neural networks \cite{pathak2016context,iizuka2017globally,yu2018generative,Liu_2018_ECCV,Zheng_2019_CVPR,Liu2019MEDFE,Nazeri_2019_ICCV,yi2020contextual,peng2021generating,zeng2021generative,Wan_2021_ICCV}.

A main challenge in this task is the requirement of \emph{bridging and exploiting visible information globally, after it had been degraded by arbitrary masks}. As depicted on the left of Fig.~\ref{fig:example}, when the entire person is masked, the natural expectation is to complete the masked area based on the visible background context. In contrast, on the right of Fig.~\ref{fig:example}, when large free-form irregular masks cover the main parts but leave the partial information visible, it is necessary but highly challenging to correctly capture \emph{long-range} dependencies between the separated foreground regions, so that the masked area can be completed in not just a photorealistic, but also semantically correct, manner.

To achieve this goal, several \emph{two-stage} approaches \cite{yu2018generative,Nazeri_2019_ICCV,yi2020contextual,peng2021generating,zeng2021generative} have been proposed, consisting of a \emph{content inference network} and an \emph{appearance refinement network}. They typically infer a coarse image/edge/semantic map based on globally visible context in a first phase, and then fill in visually realistic appearance in a second phase. However, this global perception is achieved by repeating \emph{local} convolutional operations, which have several limitations. First, due to the translation equivariance, the information flow tends to be predominantly local, with global information only shared gradually through \emph{heat-like propagation} across multiple layers. Second, during inference, the elements between adjacent layers are connected via learned but \emph{fixed} weights, rather than input-dependent adaptive weightings. These issues mean long-distance messages are only delivered inefficiently in a very deep layer, resulting in a strong inclination for the network to fill holes based on nearby rather than distant visible pixels (\cf Fig.~\ref{fig:example}). 

In this paper, we propose an alternative perspective by treating image completion as a \emph{directionless sequence-to-sequence} prediction task. In particular, instead of modeling the global context using deeply stacked convolutional layers, we design a new content inference model, called TFill, that uses a \textbf{T}ransformer-based architecture to \textbf{Fill} reasonable content into the missing holes. An important insight here is that a transformer directly exploits long-range dependencies at every encoder layer through the attention mechanism, which \emph{creates an equal flowing opportunity for all visible pixels, regardless of their relative spatial positions} (Fig.~\ref{fig:conv_vs_tans} (c)). This reduces the proximity-dominant influence that can lead to semantically incoherent results.

However, it remains a challenge to directly apply these transformer models to visual generation tasks. In particular, unlike in NLP where each word is naturally treated as a vector for token embedding \cite{Vaswani_NIPS2017_attention,devlin2018bert,radford2018improving,radford2019language}, it is unclear \emph{what a good token representation should be for a visual task}. If we use every pixel as a token, the memory cost will make this infeasible except for very small \emph{downsampled} images \cite{chen2020generative,Wan_2021_ICCV}. To mitigate this issue, our model embeds the masked image into an intermediate latent space for token representation, an approach also broadly taken by recent vision transformer models \cite{carion2020end,zhu2020deformable,esser2020taming,SETR,xiao2021early}. However, unlike these models that use traditional CNN-based encoders to embed the tokens, \emph{without considering the visible information flow in image completion}, we propose a \emph{restrictive CNN} for token representation, which has a profound influence on how the visible information is connected in the network. To do so, we ensure the individual tokens represent visible information independently, each within a \emph{small} and \emph{non-overlapping} patch. This forces \emph{the long-range context relationships between tokens to be explicitly and co-equally perceived in every transformer encoder layer}, without neighboring tokens being entangled by implicit correlation through overlapping RF. As a result, each masked pixel will \emph{not} be gradually affected by neighboring visible pixels. 

\begin{figure*}[tb!]
    \centering
    \includegraphics[width=\linewidth]{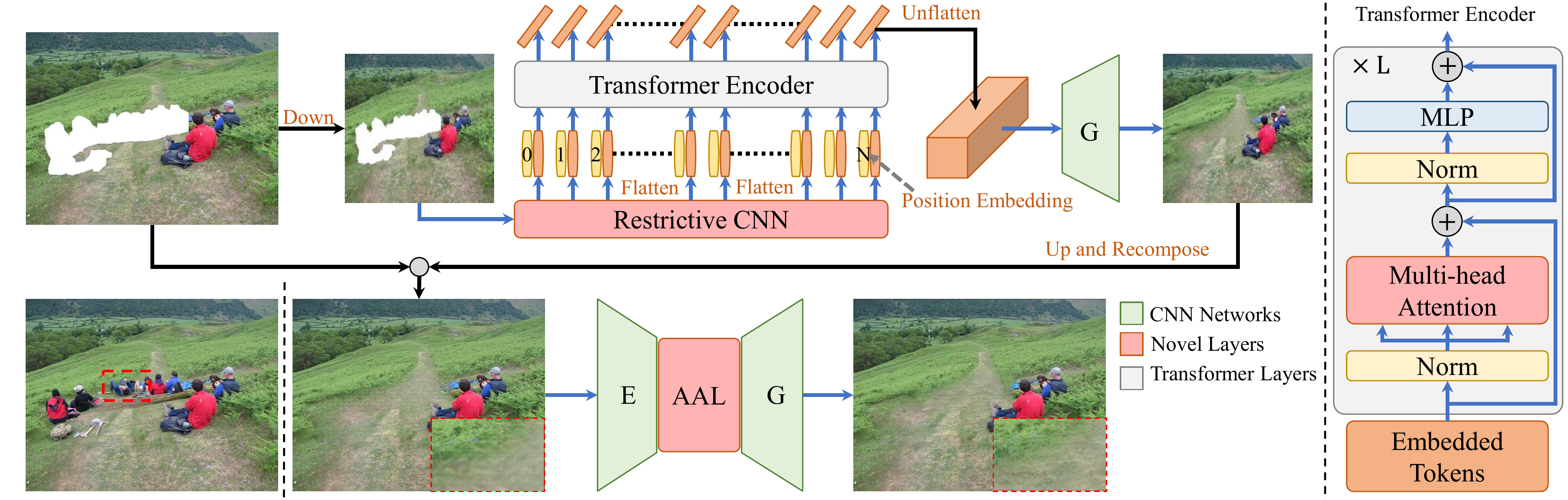}
    \begin{picture}(0,0)
    \put(-224,163){\footnotesize Masked input}
    \put(-136, 156){\tiny Fixed input ($256^2$)}
    \put(122, 156){\tiny Coarse output ($256^2$)}
    \put(-248,103){\rotatebox{90}{\footnotesize (a) Infer content}}
    \put(-248,9){\rotatebox{90}{\footnotesize (b) Refine appearance}}
    \put(-234, 7){\footnotesize Original ground truth}
    \put(-145, 7){\footnotesize Recomposed input}
    \put(40, 7){\footnotesize Refined output}
    \end{picture}
    \vspace{-0.2cm}
    \caption{\textbf{The overall pipeline of our proposed method} (The \textcolor[RGB]{255,179,179}{LightPink} boxes highlight our designed novel layers). (a) Masked input is resized to a fixed low resolution ($256\times256$) and it is then fed into a transformer encoder to generate semantically correct content. (b) The inferred content is merged with the original high-resolution image and passed to a refinement network with an \textbf{A}ttention-\textbf{A}ware \textbf{L}ayer (\textbf{AAL}) to transfer high-quality information from both visible and generated regions. Note the recomposed input has repeating artifacts, which are resolved in our refined network. Zoom in to see the details.}
    \label{fig:framework}
    \vspace{-0.2cm}
\end{figure*}
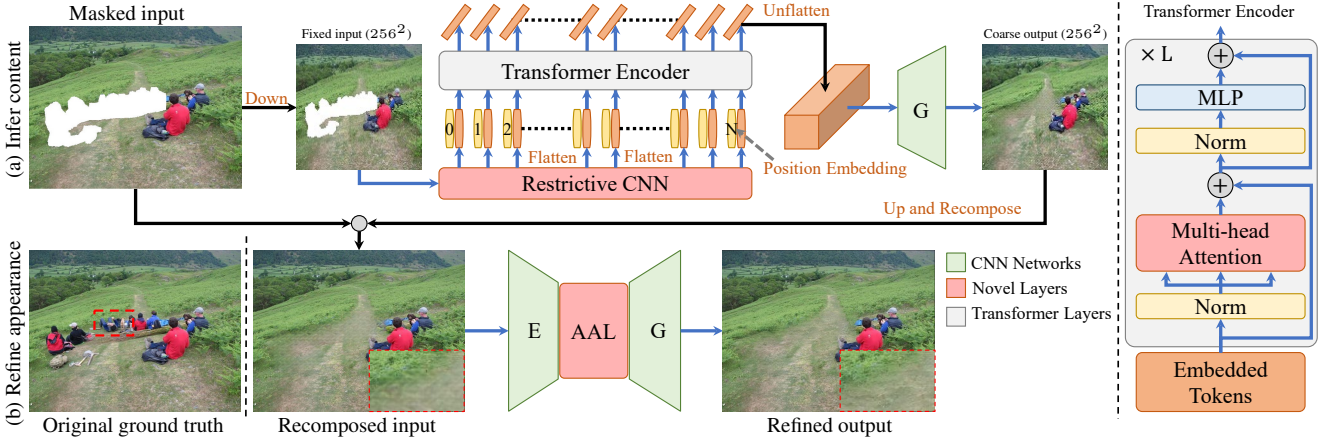

While the proposed transformer-based architecture can achieve better results than state-of-the-art methods \cite{yu2018generative,Zheng_2019_CVPR,yi2020contextual,esser2020taming}, by itself it only works for a \emph{fixed} sequence length because of the position embedding (Fig.~\ref{fig:framework}(a)). To allow our approach to flexibly scale to images of arbitrary sizes, \emph{especially at high resolution}, a fully convolutional network (Fig.~\ref{fig:framework}(b)) is subsequently applied to refine the visual appearance, building upon the coarse content previously inferred. A novel \textbf{A}ttention-\textbf{A}ware \textbf{L}ayer (AAL) is inserted between the encoder and decoder that adaptively balances the attention paid to visible and generated content, leading to semantically superior feature transfer (Figs.\ \ref{fig:attn-example} and \ref{fig:attn-comp}). 

We highlight our main contributions as follows: \textbf{1)} A \emph{restrictive} CNN head is introduced for individual \emph{weighted} token representation, which mitigates the proximity influence when propagating from local visible regions to missing holes. \textbf{2)} Through a transformer-based architecture, the long-range interactions between these tokens are explicitly modeled,
in which the masked tokens are perceptive of other visible tokens with equal opportunity, regardless of their positions. This results in a significant improvement over state-of-the-art methods. \textbf{3)} A novel attention-aware layer with adaptive attention balancing is introduced in a refined stage to obtain higher quality and resolution results. \textbf{4)} Finally, extensive experiments demonstrate that the proposed model outperforms the existing state-of-the-art image completion models.

\section{Related Work}
\label{sec:rel}

\paragraph{Image Completion:} Traditional image completion (also known as ``image inpainting'' \cite{bertalmio2000image}) methods, like diffusion-based \cite{ballester2001filling,levin2003learning,bertalmio2003simultaneous} and patch-based \cite{criminisi2004region,jia2004inference,barnes2009patchmatch}, mainly focus on background completion, by directly copying and propagating the background pixels to masked regions. 

Driven by the advances of GANs \cite{goodfellow2014generative}, CGANs \cite{mirza2014conditional} and VAEs \cite{Kingma2014}, a series of CNN-based methods \cite{pathak2016context,iizuka2017globally,yu2018generative,Liu_2018_ECCV,Zheng_2019_CVPR,Nazeri_2019_ICCV} have been proposed to hallucinate semantic meaningful content. In particular, Pathak \etal \cite{pathak2016context} introduced adversarial learning into image completion to generate new content for large holes. Iizuka \etal \cite{iizuka2017globally} extended \cite{pathak2016context}  to random regular mask. Yu \etal \cite{yu2018generative} combined the traditional patch-based idea into learning-based architecture, which is followed by \cite{song2018contextual,song2018spg,Zheng_2019_CVPR,yi2020contextual,zeng2020high,zeng2021generative}. Liu \etal \cite{Liu_2018_ECCV} proposed the partial-conv to handle random irregular masks. Zheng \etal introduced a pluralistic image completion task, aiming to generate multiple and diverse results, which is followed by \cite{zhao2020uctgan,peng2021generating,liu2021pd,Wan_2021_ICCV}. Nazeri \etal \cite{Nazeri_2019_ICCV} brought the auxiliary edge information for image completion. Then, more auxiliary information were combined into image completion in the latest Faceshape \cite{portenier2018faceshop}, DeepFill v2 \cite{yu2019free}, SC-FEGAN \cite{jo2019sc}, SWAP \cite{liao2021image}, and MST \cite{Cao_2021_ICCV}. Most of these models are built upon on a similar CNN-based encoder-decoder architecture, in which the masked regions are gradually affected by the neighboring visible pixels. Our model will solve this problem by utilizing a transformer to directly model the global context dependencies. 

\begin{figure}[tb!]
    \centering
    \includegraphics[width=\linewidth]{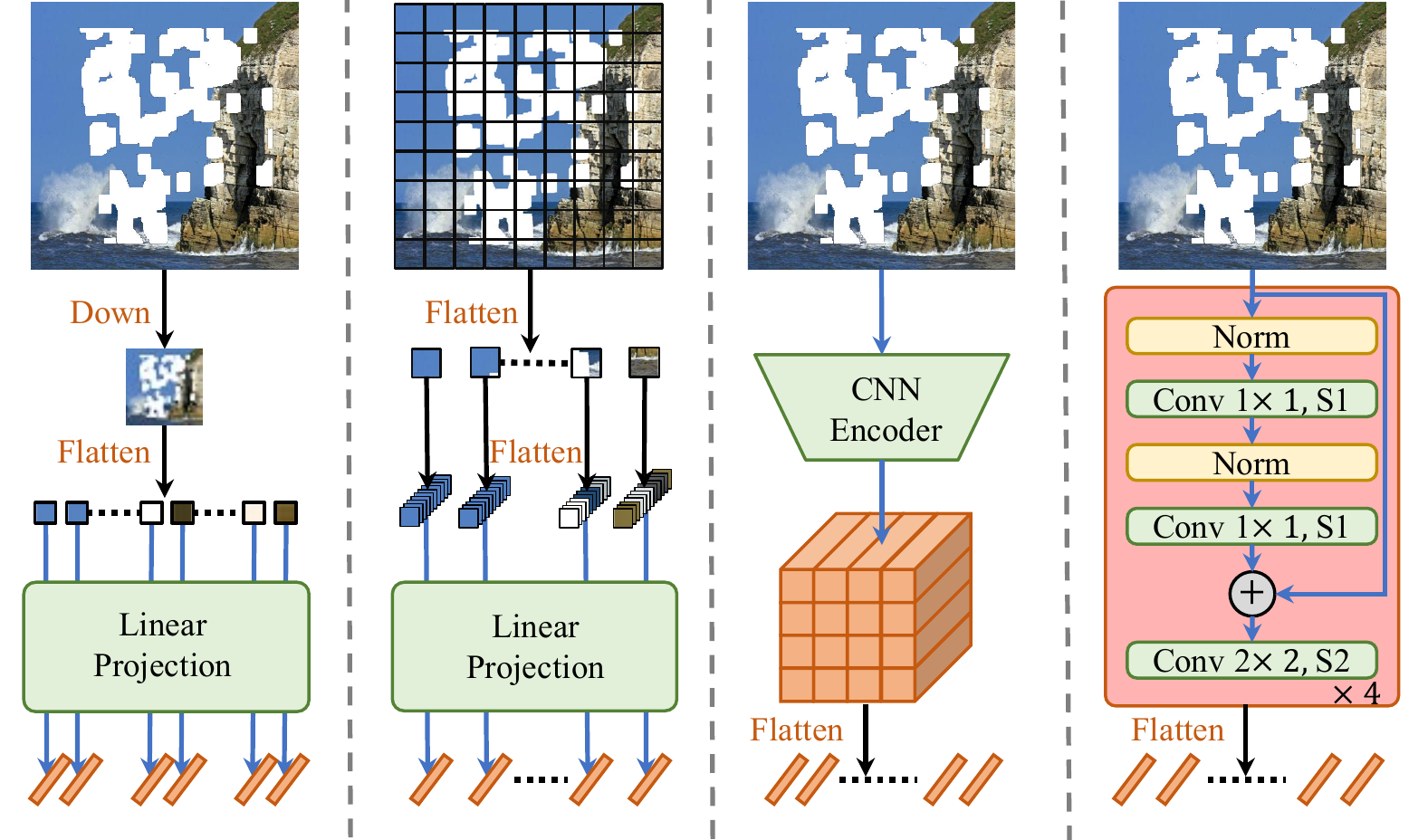}
    \begin{picture}(0,0)
    \put(-109, 9){\footnotesize (a) iGPT~\cite{chen2020generative}}
    \put(-46, 9){\footnotesize (b) VIT~\cite{dosovitskiy2020image}}
    \put(4, 9){\footnotesize (c) VQGAN~\cite{esser2020taming}}
    \put(76, 9){\footnotesize (d) Ours}
    \end{picture}
    \vspace{-0.3cm}
    \caption{\textbf{Token representation.} (a) Pixel to token. (b) Patch to token. (c) Feature to token. (d) Restrictive \textbf{R}eceptive \textbf{F}ield (RF) feature to token. Note our token has a small and non-overlapping RF like VIT \cite{dosovitskiy2020image}, but uses a stacked (x4) CNN embedding. Each token represents locally isolated patches, leaving the global context relationship to be cleanly modeled in a transformer encoder. }
    \label{fig:tokens}
\end{figure}
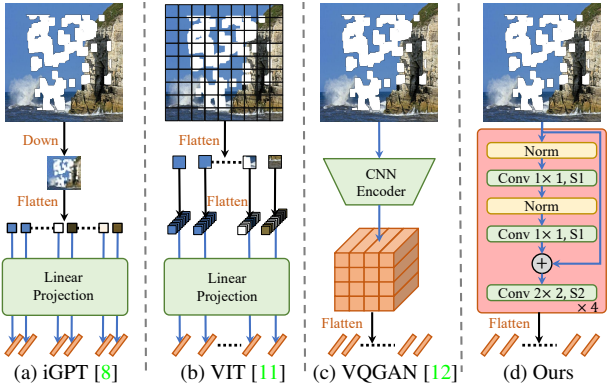

\vspace{-0.2cm}\paragraph{Visual Transformer:} The Transformer was firstly proposed by Vaswani \etal \cite{Vaswani_NIPS2017_attention} for machine translation. Inspired by the dramatic success of transformers in NLP \cite{devlin2018bert,radford2019language}, recent works have explored applying a standard transformer for vision tasks \cite{liu2021Swin}, such as image classification \cite{chen2020generative,dosovitskiy2020image,he2021masked}, object detection \cite{carion2020end,zhu2020deformable}, semantic segmentation \cite{wu2020visual,SETR}, image generation and translation \cite{esser2020taming,chen2020generative,hudson2021gansformer,jiang2021transgan}, and completion \cite{Liu_2021_FuseFormer,Wan_2021_ICCV}. Many of these embed tokens using methods shown in Fig.~\ref{fig:tokens}(a)-(c), without considering the specific information flow in image completion. In contrast, our \emph{restrictive CNN} is particularly well suited due to its compact representation in the form of local patches. 

\vspace{-0.2cm}\paragraph{Context Attention:} Context attention \cite{yu2018generative} is a specific cross-attention that aims to copy high-frequency details from high-resolution encoded features to generate high-quality images. It has recently been widely applied in image completion  \cite{yu2018generative,song2018contextual,yan2018shift,Zheng_2019_CVPR,yi2020contextual,zeng2021generative}. However, the existing works mainly copy from visible regions \cite{yu2018generative,song2018contextual,yan2018shift,yi2020contextual,zeng2021generative}, which 
is not possible for newly generated content. In addition, our AAL automatically selects features from both ``visible'' encoded and ``missing'' generated features, instead of selecting through \emph{fixed} weights \cite{Zheng_2019_CVPR}. 

\section{Methods}
\label{sec:method}

Given a masked image $\textbf{I}_m$, degraded from a real image $\textbf{I}$ by masks, our goal is to learn a model $\Phi$ to infer semantically reasonable content for missing regions, as well as filling in with visually realistic appearance. 

To achieve this, our image completion framework, illustrated in Fig.~\ref{fig:framework}, consists of a content inference network (TFill-\emph{Coarse}, Fig.~\ref{fig:framework}(a)) and an appearance refinement network (TFill-\emph{Refined}, Fig.~\ref{fig:framework}(b)). The former is responsible for capturing the global context through a transformer encoder. The embedded tokens have small receptive fields (RF) and limited capacity, \emph{preventing masked pixels' states from being implicitly dominated by visible pixels nearby than far}. While similar transformer-based architectures have recently been explored for visual tasks \cite{chen2020generative,dosovitskiy2020image,carion2020end,zhu2020deformable,esser2020taming,wu2020visual,chen2020pre,SETR,Wan_2021_ICCV}, we discover \emph{how the token representation has a profound effect on the flow of visible information in image completion, in spite of the supposedly global reach of transformers}. The latter network is designed to refine appearance by utilizing high-resolution visible features globally, and also frees the limitation to fixed sizes.

\subsection{Content Inference Network: TFill-\emph{Coarse}}
\label{sec:framework}
The proposed TFill-\emph{Coarse} model depends on the \texttt{self-attention module} in a transformer-encoder to \emph{equally} perceiving global visible context for the completed content generation. Considering the \emph{fixed} length position embedding and dramatically increased computational cost, we first downsample images with arbitrary sizes to a \emph{fixed} size, \eg $256\times256$. However, it is still \emph{not} feasible to run a transformer model if we directly \emph{flatten} image pixels into a 2D sequence with $256\times256\times3$ tokens.

To obtain a practicable number of visual tokens, different embedding methods (Fig.\ \ref{fig:tokens}(a)-(c)) have been used in current visual transformer-based works \cite{dosovitskiy2020image,carion2020end,zhu2020deformable,esser2020taming,chen2020generative,hudson2021gansformer,jiang2021transgan,wu2020visual,SETR,Wan_2021_ICCV}. These visual tokens' RF is either as small as a pixel (\eg iGPT \cite{chen2020generative}) that loses important context details due to the large-scale downsampling, or is as large as the full image size (\eg VQGAN \cite{esser2020taming}) that has firstly been gradually influenced by neighboring pixels in deep CNN layers. While patch embedding \cite{dosovitskiy2020image} achieves impressive performance in many tasks, one-layer linear projection is still not good enough \cite{xiao2021early}. The detail comparisons are reported in Table \ref{tab:ablation_transform} and Fig.\ \ref{fig:center-face-comp}.  

\vspace{-0.2cm}\paragraph{Restrictive CNN:} In contrast to these methods, our token representation is extracted using a \emph{restrictive CNN} (Fig.~\ref{fig:tokens}(d)) in 4 blocks. In each block, the $1\times1$ filter and \texttt{layernorm} is applied for non-linear projection, followed by a partial convolution layer \cite{Liu_2018_ECCV} that uses a $2$$\times$$2$ filter with stride $2$ to extract visible information. In particular, if half of the regions in a window are masked, we only embed the other $50\%$ comprising visible pixels as our token representation, and establish an initial weight of $0.5$ for the next \emph{weighted} self-attention layer. To do this, we ensure each token represents only the visible information in a local patch, \emph{leaving the long-range dependencies to be explicitly modeled by a transformer}, without cross-contamination from implicit correlation due to larger CNN RF. 

\begin{figure}[tb!]
    \centering
    \includegraphics[width=\linewidth]{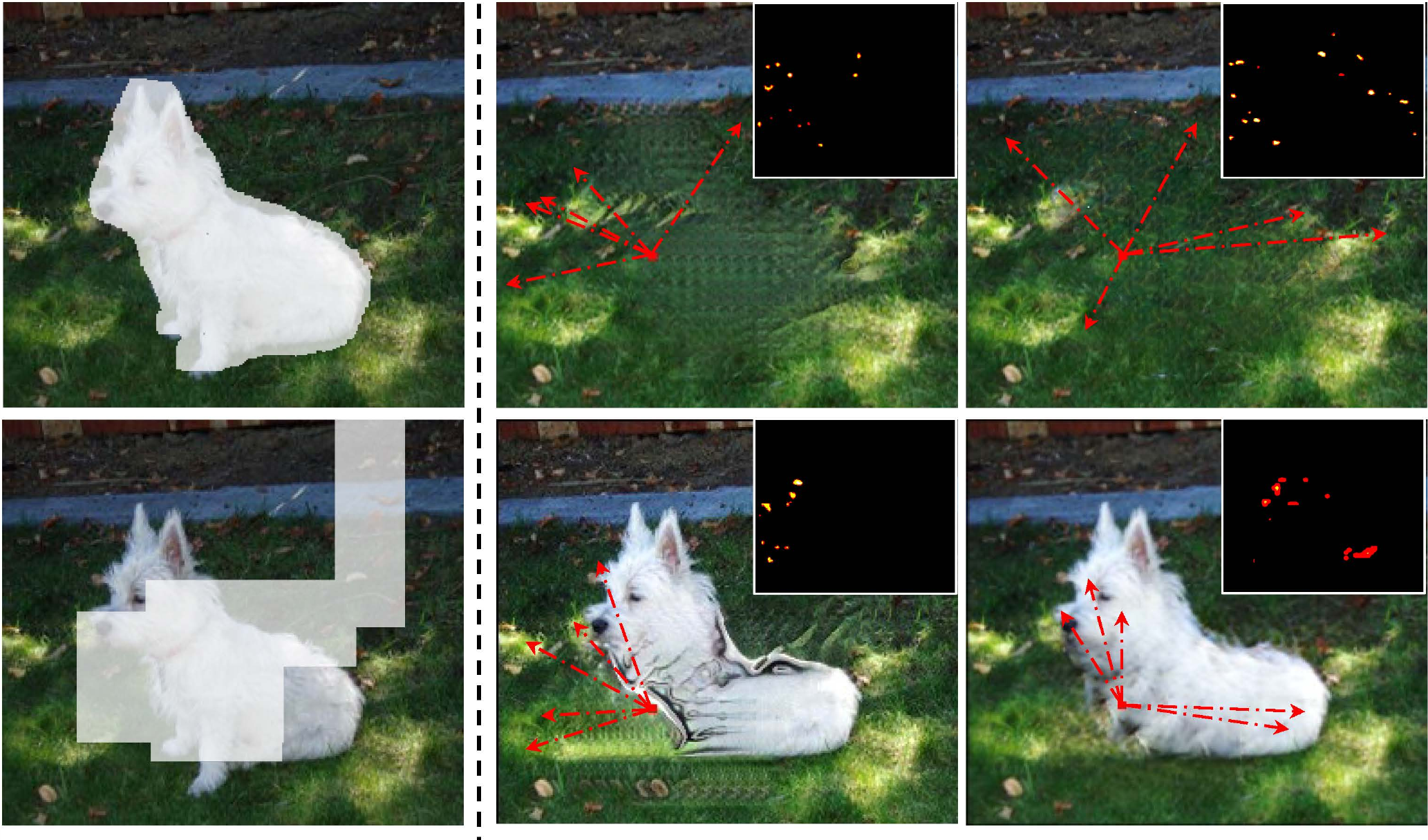}
    \begin{picture}(0,0)
    \put(-10,100){\footnotesize {\color{red}$\textbf{x}_i$}}
    \put(-10,25){\footnotesize {\color{red}$\textbf{x}_i$}}
    \put(68,100){\footnotesize {\color{red}$\textbf{x}_i$}}
    \put(68,25){\footnotesize {\color{red}$\textbf{x}_i$}}
    \put(-108,6){\footnotesize (a) Masked input}
    \put(-36,6){\footnotesize (b) HiFill~\cite{yi2020contextual}$_{\text{\scriptsize{CVPR'2020}}}$}
    \put(56,6){\footnotesize (c) Ours TFill}
    \end{picture}
    \vspace{-0.3cm}
    \caption{\textbf{An example of information flow in image completion.} The position $\textbf{x}_i$'s response (flow) is calculated by inferring the \emph{Jacobian} matrix between it to all pixels in the given masked input. Here, only the highest flows are shown. Our TFill correctly captures long-range visible context flow, even with a large mask splitting two semantically important zones.}
    \label{fig:conv_vs_tans}
\end{figure}
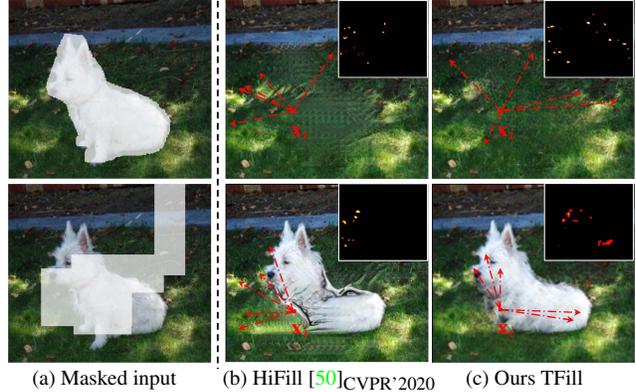

In fact, some latest works also begin to explore the influence of different token embeddings. Swin \cite{liu2021Swin} used shift windows to get multi-scales embedded features for multi-scales transformer blocks. $\text{ViT}_c$ \cite{xiao2021early} demonstrated an early CNN token embedding is important for visual transformer. However, they do not consider information flowing from visible to masked regions. When a large RF is applied into a deep CNN embedding, the masked holes will be gradually determined by the neighboring visible pixels. In Fig.\ \ref{fig:conv_vs_tans}, we empirically show this is precisely the case for prior CNN-based models. Because masked regions originally hold zero values, they will take the neighboring visible pixels as a filled and reasonable value for the next layer. In contrast, as the small patch is directly embedded using local visible information with important \emph{weight}, the proposed \emph{restrictive} CNN is better suited for image completion task. 

\vspace{-0.2cm}\paragraph{\emph{Weighted} Self-Attention Layer:} To further encourage the model to \emph{bias} to the important visible values, we replace the self-attention layer with a \emph{weighted} self-attention layer, in which a weight is applied to scale the attention scores. The initial weight $w^{(1)}\in(0.02,1.0]$ is obtained by calculating the fraction of visible pixels in a small patch, \eg $192/(16\times16)$ means $192$ pixels in the $16\times16$ patch are visible. It will then be gradually amplified by updating $w^{(i+1)}\gets\sqrt{w^{(i)}}$ after every encoder layer, to \emph{reflect} visible information flow. This initial ratio for each token is efficiently implemented in our restrictive CNN encoder. The implementation details can be found in Appendix \ref{app:sec:msa}. 

\vspace{-0.2cm}\paragraph{CNN-based Decoder:} Following existing works \cite{yu2018generative,Zheng_2019_CVPR,yi2020contextual}, a gradual upsampling decoder is implemented to generate photorealistic images. Instead of generating tokens one-by-one, our model directly predict all tokens in one step, resulting in a much faster testing time than existing transformer-based generation networks \cite{chen2020generative,esser2020taming,Wan_2021_ICCV} (Table \ref{tab:ablation_transform}).

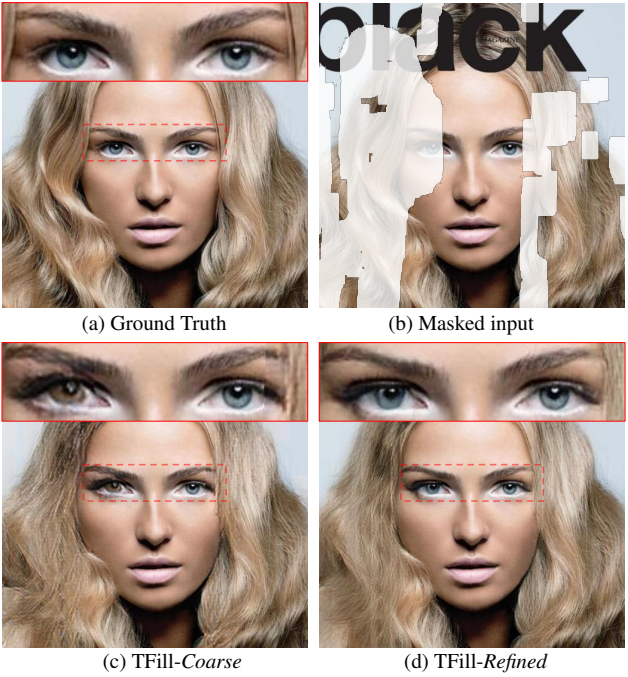
\begin{figure}[tb!]
    \centering
    \includegraphics[width=\linewidth]{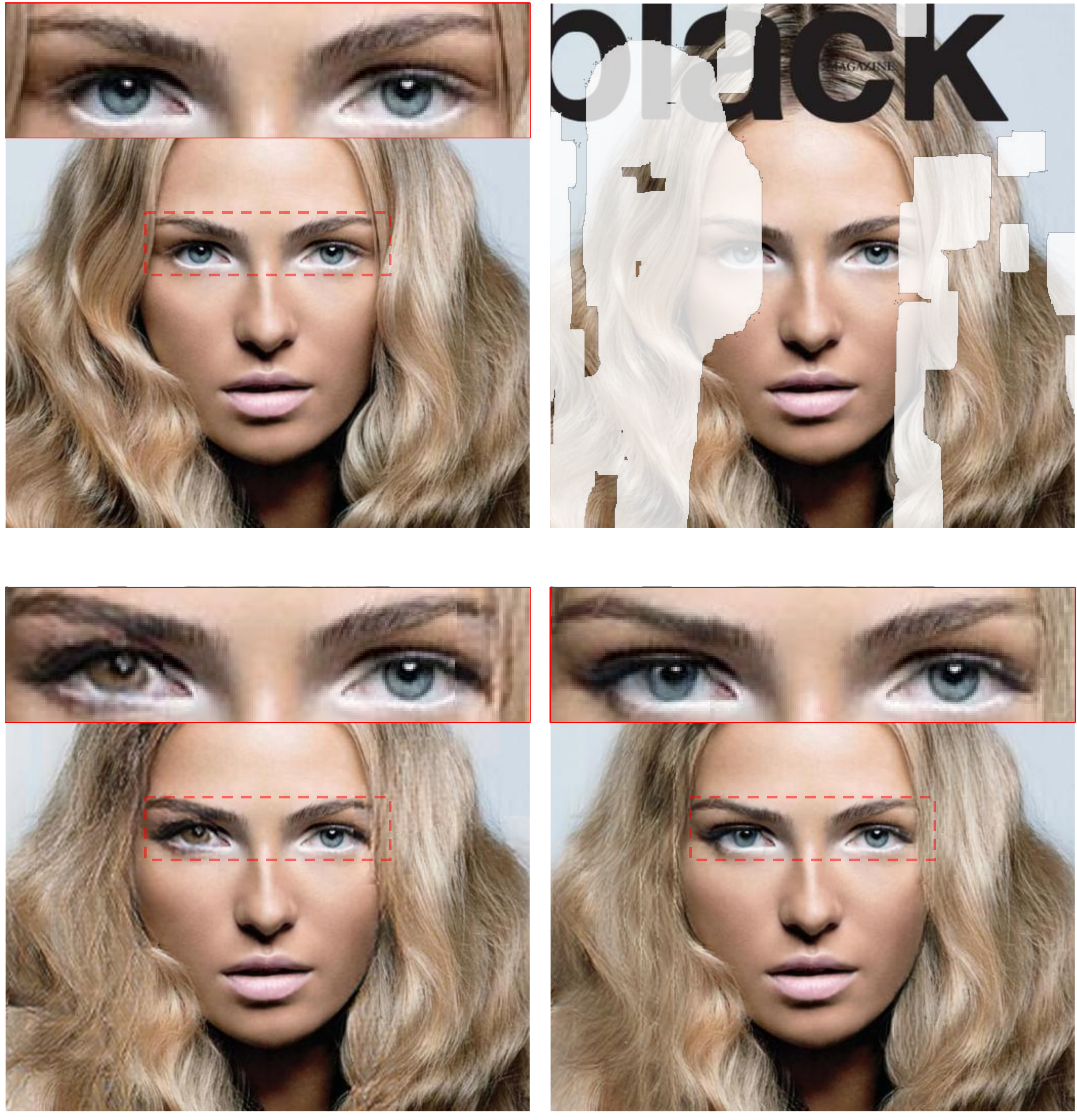}
    \begin{picture}(0,0)
    \put(-86, 134){\footnotesize (a) Ground Truth}
    \put(30, 134){\footnotesize (b) Masked input}
    \put(-78, 6){\footnotesize (c) TFill-\emph{Coarse}}
    \put(36, 6){\footnotesize (d) TFill-\emph{Refined}}
    \end{picture}
    \vspace{-0.3cm}
    \caption{\textbf{Coarse and Refined results}. (a) Ground truth. (b) Masked input degraded by free-form masks. (c) Coarse output. (d) Refined output. We can see that the refinement network not only increases image quality to a high resolution ($256^2$ \emph{vs} $512^2$), but also encourages the left eyeball to be consistent with the visible right eyeball using our attention-aware layer.}
    \label{fig:attn-example}
\end{figure}

\subsection{Appearance Refinement Network: TFill-\emph{Refined}}
\label{sec:attn-aware}

Although the proposed TFill-\emph{Coarse} model correctly infers superiorly reasonable content (shown in Figs.\ \ref{fig:sup-center-imgnet-examples}, \ref{fig:sup-center-face-examples}, and \ref{fig:sup-center-places2-examples}) by equally utilizing the global visible context in every layer, two limitations remain. First, it is \emph{not} suitable for high-resolution input due to the \emph{fixed} length position embedding. Second, the realistic completed results may \emph{not be fully consistent with the original visible appearances}, \eg the generated left eye having a different shape and color to the visible right eye in Fig.~\ref{fig:attn-example} (c). 

\vspace{-0.2cm}\paragraph{Attention-Aware Layer (AAL):}  To mitigate these issues, a refinement network, trained on high-resolution images, is proposed (Fig.~\ref{fig:framework} (b)). In particular, to further utilize the visible high-frequency details in global, an \textbf{A}ttention-\textbf{A}ware \textbf{L}ayer (AAL) is designed to \emph{copy long-range information from both encoded and decoded features}. 

\begin{figure}[tb!]
    \centering
    \includegraphics[width=\linewidth]{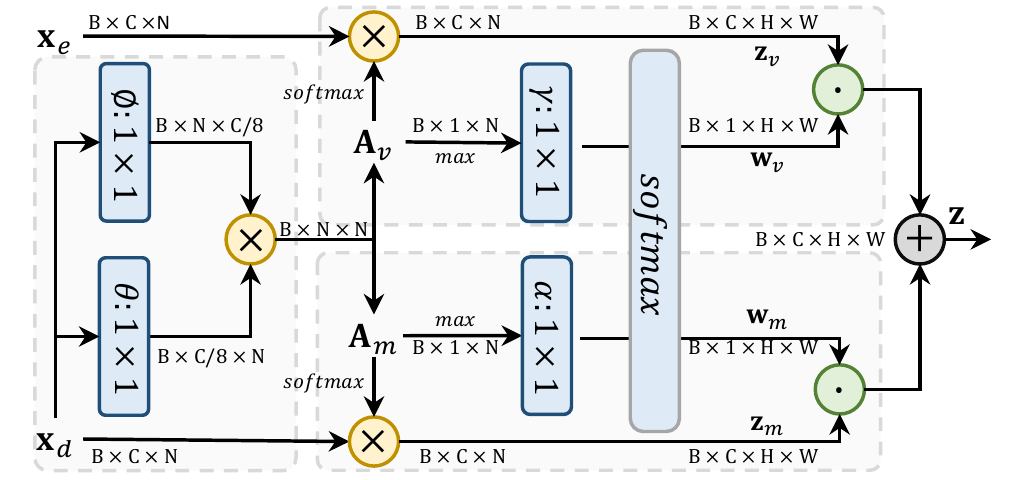}
    \vspace{-0.4cm}
    \caption{\textbf{Attention-aware layer.} The feature maps are shown as tensors. ``$\bigotimes$'' denotes matrix multiplication, ``$\bigodot$'' denotes element-wise multiplication and ``$\bigoplus$'' is element-wise sum. The blue boxes denote $1\times1$ convolution filters that are learned.}
    \label{fig:attn-layer}
\end{figure}

As depicted in Fig.~\ref{fig:attn-layer}, given a decoded feature ${\bf x}_d$, we first calculate the attention score of:
\begin{equation}
    {\bf A} = \phi({\bf x}_d)^\intercal \theta({\bf x}_d),
\end{equation}
where ${\bf A}_{ij}$ represents the similarity of the $i^{\text{th}}$ feature to the $j^{\text{th}}$ feature, and $\mathbf{\phi}$, $\mathbf{\theta}$ are $1$$\times$$1$ convolution filters.

Interestingly, we discover that using ${\bf A}$ directly in a standard self-attention layer is suboptimal, because the ${\bf x}_d$ features for visible regions are generally distinct from those generated for masked regions. Consequently, \emph{the attention tends to be insular}, with masked regions preferentially attending to masked regions, and vice versa. To avoid this problem, we explicitly handled the attention to visible regions separately from masked regions. So before \texttt{softmax} normalization, ${\bf A}$ is split into two parts: ${\bf A}_v$ --- similarity to \emph{visible} regions, and ${\bf A}_m$ --- similarity to generated \emph{masked} regions. Next, we get long-range dependencies via:
\begin{equation}
    \begin{matrix}
    {\bf z}_v = \texttt{\small softmax}({\bf A}_v){\bf x}_e
    & ,
    & {\bf z}_m = \texttt{\small softmax}({\bf A}_m){\bf x}_d
    \end{matrix}
\end{equation}
where ${\bf z}_v$ contains features of contextual flow \cite{yu2018generative} for copying high-frequency details from the encoded high-resolution features ${\bf x}_e$ to masked regions, while ${\bf z}_m$ has features from the self-attention that is used in SAGAN \cite{zhang2019self} for high-quality image generation. 

Instead of learning \emph{fixed} weights \cite{Zheng_2019_CVPR} to combine ${\bf z}_v$ and ${\bf z}_m$, we learn the \emph{weights mapping} based on the largest attention score in each position. Specifically, we first obtain the largest attention score of ${\bf A}_v$ and ${\bf A}_m$, respectively. Then, we use the $1$$\times$$1$ filter $\gamma$ and $\alpha$ to \emph{modulate} the ratio of the weights. \texttt{Softmax} normalization is applied to ensure ${\bf w}_v$$+$${\bf w}_m$$=$$1$ in every spatial position:
\begin{equation}
    [{\bf w}_v, {\bf w}_w] = \texttt{\small softmax}([\gamma(\texttt{\small max}({\bf A}_v)),\alpha(\texttt{\small max}({\bf A}_m)]))
\end{equation}
where \texttt{\small max} is executed on the attention score channel. Finally,  an attention-balanced output ${\bf z}$ is obtained by:
\begin{equation}
    {\bf z} = {\bf w}_v \cdot {\bf z}_v +  {\bf w}_m \cdot {\bf z}_m
\end{equation}
where ${\bf w}_v, {\bf w}_m \in\mathbb{R}^{B\times 1\times H\times W}$ hold different values for various positions, dependent on the largest attention scores in the visible and masked regions, respectively. 

\section{Experiments}

\begin{figure*}[tb!]
    \centering
    \includegraphics[width=\linewidth]{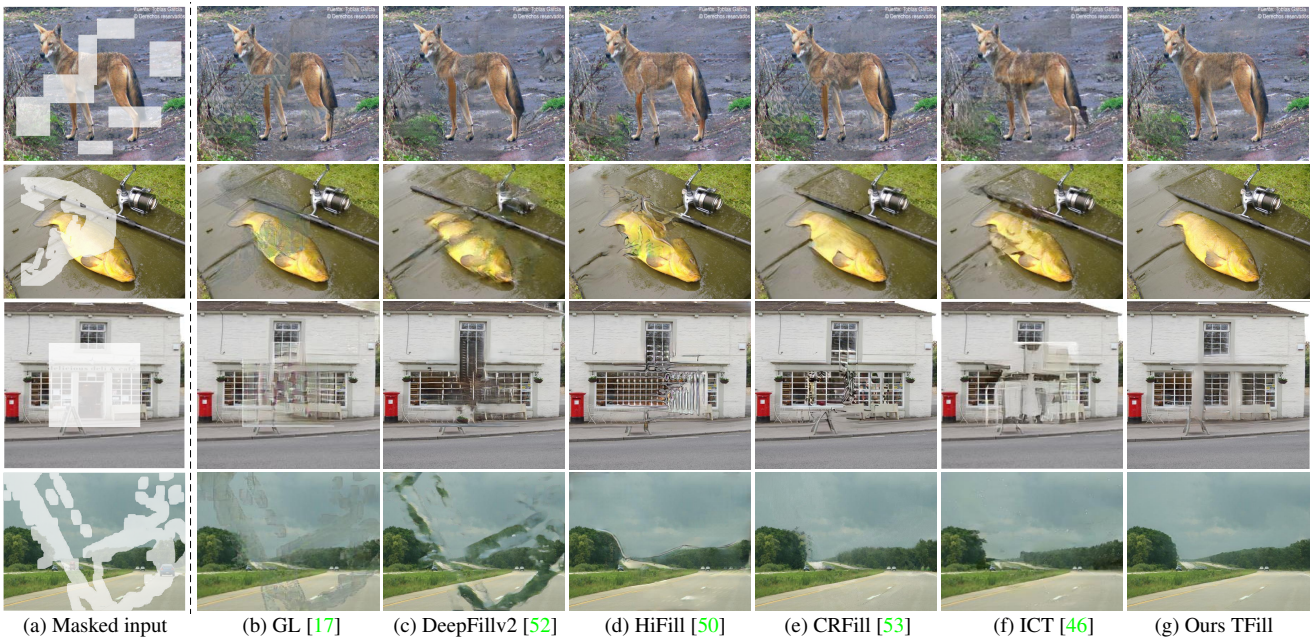}
    \begin{picture}(0,0)
    \put(-238, 5){\footnotesize (a) Masked input}
    \put(-155, 5){\footnotesize (b) GL~\cite{iizuka2017globally}}
    \put(-98, 5){\footnotesize (c) DeepFillv2~\cite{yu2019free}}
    \put(-18, 5){\footnotesize (d) HiFill~\cite{yi2020contextual}}
    \put(50, 5){\footnotesize (e) CRFill~\cite{zeng2021generative}}
    \put(129, 5){\footnotesize (f) ICT~\cite{Wan_2021_ICCV}}
    \put(190, 5){\footnotesize (g) Ours TFill}
    \end{picture}
    \vspace{-0.3cm}
    \caption{\textbf{Qualitative comparison on various datasets with free-form masks.} Here, we show results for ImageNet \cite{russakovsky2015imagenet} (top two examples) and Places2 \cite{zhou2018places} (bottom two examples). Our model generated more reasonable object and scene  structures, with better visual results. Please zoom in to see the details. More comparisons are provided in Figs. \ref{fig:sup-free-form-face}, \ref{fig:sup-free-form-imagenet}, and \ref{fig:sup-free-form-place}.}
    \label{fig:results_free-comp}
\end{figure*}

\begin{table*}[tb!]
    \centering
    \footnotesize
    \renewcommand{\arraystretch}{1.1}
    \setlength\tabcolsep{3pt}
    \begin{tabular}{@{}lccccccccccccccc@{}}
        \hlineB{3.5}
         & \multicolumn{3}{c}{PSNR$\uparrow$} && \multicolumn{3}{c}{SSIM$\uparrow$} && \multicolumn{3}{c}{LPIPS$\downarrow$} && \multicolumn{3}{c}{FID$\downarrow$} \\
         \cline{2-4}\cline{6-8}\cline{10-12}\cline{14-16} 
        Mask Ratio & 20-30\% & 30-40\% & 40-50\% && 20-30\% & 30-40\% & 40-50\% && 20-30\% & 30-40\% & 40-50\% && 20-30\% & 30-40\% & 40-50\% \\
        \hlineB{2.5}
        GL \cite{iizuka2017globally}$_{\text{\scriptsize{SIGGRAPH'2017}}}$ &  21.33 & 19.11 & 17.56 && 0.7672 & 0.6823 & 0.5987 && 0.1847 & 0.2535 & 0.3189 && 39.22 & 53.24 & 68.46 \\
        PIC \cite{Zheng_2019_CVPR}$_{\text{\scriptsize{CVPR'2019}}}$ & 24.44 & 22.32 & 20.71 && 0.8520 & 0.7850 & 0.7119 && 0.1183 & 0.1666 & 0.2245 && 21.62 & 29.59 & 41.60 \\
        DeepFillv2 \cite{yu2019free}$_{\text{\scriptsize{ICCV'2019}}}$ & 23.58 & 21.50 & 19.94 && 0.8319 & 0.7712 & 0.7074 && 0.1234 & 0.1639 & 0.2079 && 23.18 & 28.87 & 35.21 \\
        HiFill \cite{yi2020contextual}$_{\text{\scriptsize{CVPR'2020}}}$ & 22.54 & 20.15 & 18.48 && 0.7838 & 0.7057 & 0.6193 && 0.1632 & 0.2258 & 0.3053 && 26.89 & 38.40 & 56.24 \\
         CRFill \cite{zeng2021generative}$_{\text{\scriptsize{ICCV'2021}}}$ & 24.38 &  21.95 & 20.44 && 0.8476 & 0.7983 & 0.7217 && 0.1189 & 0.1597 & 0.1993 && 17.58 & 23.05 & 29.97 \\
         ICT \cite{Wan_2021_ICCV}$_{\text{\scriptsize{ICCV'2021}}}$ & 24.53 & 22.84 & 21.11 && 0.8599 & 0.7995 & 0.7228 && 0.1045 & 0.1563 & 0.1974 && 17.13 & 22.39 & 28.18 \\
         Ours TFill & \textbf{25.10} & \textbf{22.89} & \textbf{21.22} && \textbf{0.8686} & \textbf{0.8063} & \textbf{0.7391} && \textbf{0.0918} & \textbf{0.1328} & \textbf{0.1796} && \textbf{15.28} & \textbf{19.99} & \textbf{25.88} \\
        \hlineB{3.5}
    \end{tabular}
    \vspace{-0.2cm}
    \caption{\small{Quantitative comparisons on Places2~\cite{zhou2018places} with free-form masks \cite{Liu_2018_ECCV}. Without bells and whistles, TFill outperformed all existing learning-based models. The results are reported on $256\times256$ resolution, as earlier works were trained only on this scale.}}
    \label{tab:SOTA_comp}
\end{table*}

\subsection{Experimental Details}

\paragraph{Datasets:} We evaluated the proposed TFill model with arbitrary mask types on various datasets, including CelebA-HQ \cite{liu2015faceattributes,karras2018progressive}, FFHQ \cite{karras2019style}, Places2 \cite{zhou2018places}, and ImageNet \cite{russakovsky2015imagenet}.

\vspace{-0.2cm}\paragraph{Metrics:} Following existing works \cite{Nazeri_2019_ICCV,Wan_2021_ICCV,zheng2021pluralistic}, we mainly reported the traditional patch-level image quality metrics, including peak signal-to-noise ratio (PSNR) and structure similarity index (SSIM), and the latest learned feature-level LPIPS \cite{zhang2018perceptual} and FID \cite{heusel2017gans} metrics. 

\vspace{-0.2cm}\paragraph{Implementation Details:} Our model is trained in two stages: \textbf{1)} the TFill-\emph{Coarse} is first trained for $256\times256$ resolution; and \textbf{2)} the TFill-\emph{Refined} is then trained for $512\times512$ resolution. Unless other noted, TFill indicates the whole model in the paper. Both networks are optimized using the loss $L = L_{pixel} + L_{per} + L_{GAN}$, where $L_{pixel}$ is the $\ell_1$ reconstruction loss, $L_{per}$ is the perceptual loss \cite{johnson2016perceptual}, and $L_{GAN}$ is the discriminator loss \cite{goodfellow2014generative}. More implementation details are provided in Appendix \ref{app:sec:experiment}. 

\subsection{Main Results}

We firstly compared with the following state-of-the-art image completion methods: GL~\cite{iizuka2017globally}$_{\text{\scriptsize{SIGGRAPH'2017}}}$, DeepFillv2~\cite{yu2019free}$_{\text{\scriptsize{ICCV'2019}}}$, PIC~\cite{Zheng_2019_CVPR}$_{\text{\scriptsize{CVPR'2019}}}$, HiFill~\cite{yi2020contextual}$_{\text{\scriptsize{CVPR'2020}}}$, CRFill~\cite{zeng2021generative}$_{\text{\scriptsize{ICCV'2021}}}$, and ICT~\cite{Wan_2021_ICCV}$_{\text{\scriptsize{ICCV'2021}}}$ using their publicly released codes and models. 

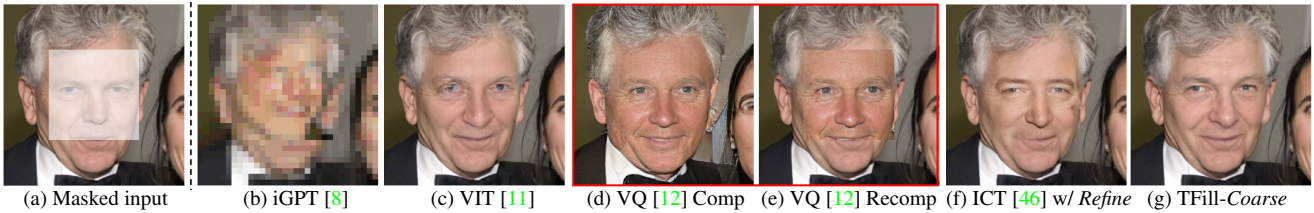
\begin{figure*}[tb!]
    \centering
    \includegraphics[width=\linewidth]{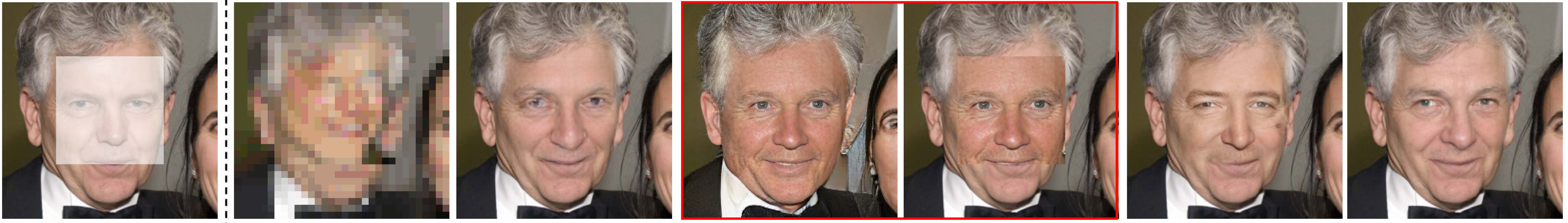}
    \begin{picture}(0,0)
    \put(-239, 6){\footnotesize (a) Masked input}
    \put(-156, 6){\footnotesize (b) iGPT~\cite{chen2020generative}}
    \put(-85, 6){\footnotesize (c) VIT~\cite{dosovitskiy2020image}}
    \put(-26, 6){\footnotesize (d) VQ~\cite{esser2020taming} Comp}
    \put(40, 6){\footnotesize (e) VQ~\cite{esser2020taming} Recomp}
    \put(110, 6){\footnotesize (f) ICT~\cite{Wan_2021_ICCV} w/ \emph{Refine}}
    \put(186, 6){\footnotesize (g) TFill-\emph{Coarse}}
    \end{picture}
    \vspace{-0.3cm}
    \caption{\textbf{Comparing results under different token representations.} All transformers are based on the same transformer backbone ~\cite{Vaswani_NIPS2017_attention}. For VQGAN \cite{esser2020taming}, we report completed (Comp) image and recomposed (Recomp) image. ICT \cite{Wan_2021_ICCV} used two-stages networks as the original paper. TFill-\emph{Coarse} is our model with configure $\mathbb{E}$ in Table \ref{tab:conv_vs_transform}, \ie TFill w/o the refinement network.}
    \vspace{-0.3cm}
    \label{fig:center-face-comp}
\end{figure*}

\vspace{-0.2cm}\paragraph{Quantitative Results:} Table \ref{tab:SOTA_comp} shows quantitative evaluation results on Places2 \cite{zhou2018places}, in which the images were degraded by free-form masks provided in the PConv \cite{Liu_2018_ECCV} testing set. The mask ratio denotes the range of masking proportion applied to the images. The original mask ratios hold six levels, from 0 to $60\%$, increasing $10\%$ for each level. Here, following ICT \cite{Wan_2021_ICCV}, we only compare the results on middle-level mask ratios. As can be seen, the proposed TFill model outperformed the CNN-based state-of-the-art models in all mask scales. Spcifically, it achieves averaging relative $18.8\%$ and $13.3\%$ improvements for LPIPS and FID scores, respectively. While the latest ICT \cite{Wan_2021_ICCV} also utilized the transformer architecture to capture the global information, they directly downsampled the original image into 32$\times$32, or $48\times48$ resolution, and then embeded each pixel as a token, resulting in important information is lost during such hard large-scale downsampling.

\begin{table}[tb!]
    \centering
    \renewcommand{\arraystretch}{1.0}
    \setlength\tabcolsep{4pt}
    \begin{tabular}{@{}llccccc@{}}
         \hlineB{3.5}
         & \multirow{2}{*}{\textbf{Method}} & \multicolumn{2}{c}{\textbf{CelebA-HQ}}&& \multicolumn{2}{c}{\textbf{FFHQ}}\\
		\cline{3-4}\cline{6-7}
		& & LPIPS$\downarrow$& FID$\downarrow$ && LPIPS$\downarrow$ & FID$\downarrow$\\
		\hlineB{2}
		\multicolumn{2}{l}{CA~\cite{yu2018generative}$_{\text{\scriptsize{CVPR'2018}}}$} & 0.104 & 9.53 &  & 0.127 & 8.78\\
		\multicolumn{2}{l}{PIC~\cite{Zheng_2019_CVPR}$_{\text{\scriptsize{CVPR'2019}}}$} & 0.061 & 6.43 & & 0.068 & 4.61\\
		\multicolumn{2}{l}{MEDFE~\cite{Liu2019MEDFE}$_{\text{\scriptsize{ECCV'2020}}}$} & 0.067 & 7.01 & & - & - \\
		\cdashline{1-7}
		$\mathbb{A}$ & Traditional \emph{Conv} & 0.060 & 6.29 && 0.066 & 4.12 \\
		$\mathbb{B}$ & + Attention in G & 0.059 & 6.34 & & 0.064 & 4.01 \\
		$\mathbb{C}$ & + Restrictive \emph{Conv} &  0.056 & 4.68 && 0.060 & 3.87 \\
		$\mathbb{D}$ & + Transformer & 0.051 & 4.02 && 0.057 & 3.66\\
		$\mathbb{E}$ & + Masked Attention &  0.050 & 3.92 && 0.057 & 3.63\\
		$\mathbb{F}$ & + Refine Network & \textbf{0.048} & \textbf{3.86} & & \textbf{0.053} & \textbf{3.50} \\
        \hlineB{2}
    \end{tabular}
    \caption{Learned Perceptual Image Patch Similarity (LPIPS) and Fr\'echet Inception Distance (FID) for various completion networks on center masked images. In this paper, we calculate the LPIPS and FID using all images in the corresponding test sets.}
    \label{tab:conv_vs_transform}
\end{table}

\vspace{-0.2cm}\paragraph{Qualitative Results:} The qualitative comparisons are visualized in Figs. \ref{fig:results_free-comp} and \ref{fig:attn-comp}. The proposed TFill achieved superior visual results even under challenging conditions. In Fig.\ \ref{fig:attn-comp}, we compared with CA \cite{yu2018generative}, PIC \cite{Zheng_2019_CVPR}, and CRFill \cite{zeng2021generative} on Celeba-HQ dataset. Our TFill generates photorealistic high-resolution ($512\times512$) results, even when significant semantic information is missing. 

Fig.\ \ref{fig:results_free-comp} shows visual results on natural images that were degraded by random masks. Here, we mainly compared the results for semantic content completion, while visualizing the easily traditional object removal results in Appendix \ref{app:sec:removing} (Figs. \ref{fig:sup-free-edit-face}, \ref{fig:sup-free-edit-imagenet0}, and \ref{fig:sup-free-edit-imagenet}). GL \cite{iizuka2017globally}, DeepFillv2 \cite{yu2019free}, and HiFill \cite{yi2020contextual}, while good at object removal, failed to infer shapes needed for object completion, \eg the content for animals. CRFill \cite{zeng2021generative} provided plausible appearance, yet the animals' shapes are unaligned, \eg malposed leg and body of the dog. Our TFill inferred the correct shapes for even heavily masked objects in ImageNet, \eg the fish even with head and tail separated by a large mask. It also outperformed all previous methods on high-resolution masked images in Places2, especially for some large masked regions. More comparisons are presented in Appendix \ref{app:sec:comparisons} (Figs. \ref{fig:sup-free-form-face}, \ref{fig:sup-free-form-imagenet}, \ref{fig:sup-free-form-place}). Please zoom in to see the details.

\subsection{Ablation Experiments}

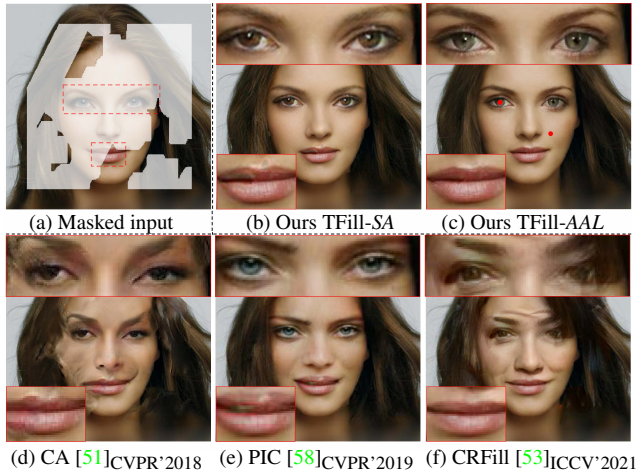
\begin{figure}[tb!]
    \centering
    \includegraphics[width=\linewidth]{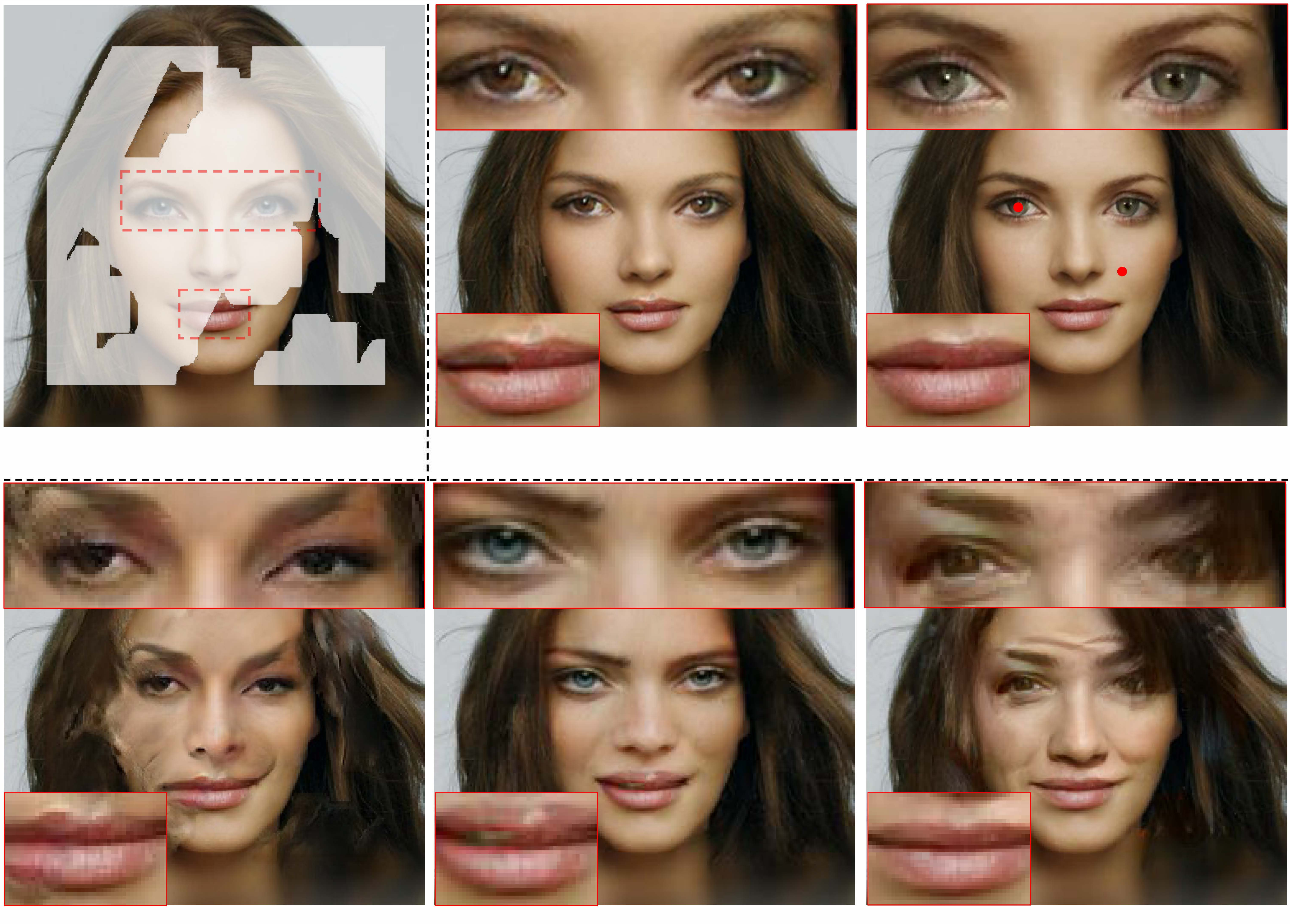}
    \begin{picture}(0,0)
    \put(-108, 96){\footnotesize (a) Masked input}
    \put(-26, 96){\footnotesize (b) Ours TFill-\emph{SA}}
    \put(48, 96){\footnotesize (c) Ours TFill-\emph{AAL}}
    \put(-115, 7){\footnotesize (d) CA~\cite{yu2018generative}$_{\text{\scriptsize{CVPR'2018}}}$}
    \put(-36, 7){\footnotesize (e) PIC~\cite{Zheng_2019_CVPR}$_{\text{\scriptsize{CVPR'2019}}}$}
    \put(42, 7){\footnotesize (f) CRFill~\cite{zeng2021generative}$_{\text{\scriptsize{ICCV'2021}}}$}
    \end{picture}
    \vspace{-0.3cm}
    \caption{\textbf{Results with different attention modules} in various methods. Our attention-aware layer is able to adaptively select the features from both visible and generated content.}
    \label{fig:attn-comp}
\end{figure}

We ran a number of ablations to analyze the effectiveness of each component in our TFill. Results are shown in Tables \ref{tab:conv_vs_transform}, \ref{tab:ablation_transform}, and \ref{tab:attn_aware_ana}, and Figs. \ref{fig:center-face-comp} and \ref{fig:attn-comp}.

\vspace{-0.2cm}\paragraph{TFill Architecture:} We first evaluated components in the redesigned image completion architecture in Table \ref{tab:conv_vs_transform}, which experimentally demonstrates that the new architecture considerably improves the performance. Our baseline configuration ($\mathbb{A}$) used an encoder-decoder structure derived from VQGAN \cite{esser2020taming}, except here attention layers were removed in advance for a pure CNN-baseline. When combined with the powerful discriminator of StyleGANv2 \cite{karras2020analyzing}, the performance was comparable to previous state-of-the-art CNN-based PIC \cite{Zheng_2019_CVPR,zheng2021pluralistic}. We first added the self-attention layer \cite{zhang2019self}, \emph{not} context mapping from the encoder \cite{yu2018generative,Zheng_2019_CVPR}, to the decoder (Generator, G) in ($\mathbb{B}$), but the performance remained similar to baseline ($\mathbb{A}$). Interestingly, when we use the proposed \emph{restrictive CNN} in ($\mathbb{C}$) to \emph{embed information in the local patch}, the performance improved substantially, especially for FID (relative $20.2\%$ on CelebA-HQ). This suggests that the input feature representation is significant for the attention layer to equally deliver all messages, as explained in Fig.\ \ref{fig:conv_vs_tans}. We then improved this new baseline by adding the transformer encoder ($\mathbb{D}$), which benefits from globally delivered messages at multipe layers. Finally, we introduced masked weights to each attention layer of the transformer ($\mathbb{E}$), improving results further.

\vspace{-0.2cm}\paragraph{Token Representation:} Tables \ref{tab:conv_vs_transform} and \ref{tab:ablation_transform} report the influence of the token representation. Our TFill achieved much better performance when using the \emph{restrictive}-CNN. iGPT \cite{chen2020generative} downsamples the image to a fixed scale, \eg $32\times32$, and embeds \emph{each pixel to a token}. While this may not impact the classification \cite{torralba200880}, it has a large negative effect on generating high-quality images. Furthermore, the autoregressive form results in the completed image being inconsistent with the bottom-right visible region (Fig.~\ref{fig:center-face-comp} (b)), and each image runs an average of 26.45s on an NVIDIA 1080Ti GPU. ICT \cite{Wan_2021_ICCV} improved iGPT by using bidirectional attention and adding a guided upsampling network. While the refined performance can almost match our coarse results, the running time is ruinously expensive (average 152.48s/img) and the content is \emph{not} aligned well in Figs. \ref{fig:example} and \ref{fig:results_free-comp}. In contrast, VIT \cite{dosovitskiy2020image} embeds \emph{each patch to a token}. As shown in Table \ref{tab:ablation_transform} and Fig.~\ref{fig:center-face-comp}, it can achieve relatively good quantitative and qualitative results. However, some details are perceptually poor, \eg the strange eyes in Fig.~\ref{fig:center-face-comp}. Finally, VQGAN \cite{esser2020taming} employs a large RF CNN to embed the image. It generates a visually realistic completion (Fig.\ \ref{fig:center-face-comp} (d)), but when pasted to the original input (Fig.\ \ref{fig:center-face-comp} (e)), there is an obvious gap between generated and visible pixels. When we used large convolutional kernels for large RF (229), the holes will firstly be filled in with neighboring visible pixels, resulting in worse results.

\begin{table}[tb!]
    \centering
     \renewcommand{\arraystretch}{1.0}
    \setlength\tabcolsep{3pt}
    \begin{tabular}{@{}llcccc@{}}
        \hlineB{3.5}
        & \textbf{Method} & LPIPS$\downarrow$& FID$\downarrow$  & Mem$\downarrow$ & Time$\downarrow$\\
        \hlineB{2}
        \multicolumn{2}{l}{IGPT~\cite{chen2020generative}$_{\text{\scriptsize{ICML'2020}}}$(RF $1$)} & 0.609 & 148.42 & 3.16 & 26.45\\
        \multicolumn{2}{l}{VIT~\cite{dosovitskiy2020image}$_{\text{\scriptsize{ICLR'2021}}}$(RF $16$)} & 0.062 & 5.09 & 1.16 & \textbf{0.167} \\
		\multicolumn{2}{l}{VQGAN~\cite{esser2020taming}$_{\text{\scriptsize{CVPR'2021}}}$} & 0.226 & 11.92 & 2.36 & 4.29\\
		\multicolumn{2}{l}{ICT~\cite{Wan_2021_ICCV}$_{\text{\scriptsize{ICCV'2021}}}$(RF $1$)} & 0.061 & 4.24 & 3.87 & 152.48 \\
		\cdashline{1-6}
        $\mathbb{E}$ & TFill-\emph{Coarse} (RF $229$) & 0.062 & 3.92 & 1.25 & 0.188\\
        $\mathbb{E}$ & TFill-\emph{Coarse} (RF $16$) & \textbf{0.057} & \textbf{3.63} & \textbf{1.15} & 0.180\\
        \hlineB{2}
    \end{tabular}
    \caption{The effect of restrictive token representation on FFHQ dataset. ``RF'' indicates the Receptive Field size. ``Mem'' denotes the memory (GB) cost during testing and ``Time'' is the testing time (s) for each center masked image. }
    \label{tab:ablation_transform}
     \vspace{-0.1cm}
\end{table}

\begin{table}[tb!]
    \centering
    \renewcommand{\arraystretch}{1.1}
    \setlength\tabcolsep{4pt}
    \begin{tabular}{@{}lccccc@{}}
        \hlineB{3.5}
        & \multicolumn{2}{c}{LPIPS$\downarrow$} && \multicolumn{2}{c}{FID$\downarrow$} \\
        \cline{2-3}\cline{5-6}
         Mask Type & center & random && center & random \\
         \hlineB{2}
         SA \cite{zhang2019self}$_{\text{\scriptsize{ICML'19}}}$ & 0.0584 & 0.0469 && 3.62 & 2.69 \\
         CA \cite{yu2018generative}$_{\text{\scriptsize{CVPR'2018}}}$ & 0.0608 & 0.0443  && 3.86 & 2.66 \\
         SLTA \cite{Zheng_2019_CVPR}$_{\text{\scriptsize{CVPR'2019}}}$ & 0.0561 & 0.0452 && 3.61 & 2.64 \\
         Ours-AAL & \bf{0.0533} & \bf{0.0412} && \bf{3.50} & \bf{2.57}\\
         \hlineB{2}
    \end{tabular}
    \caption{The effect of various attention layers on FFHQ dataset. ``center'' denotes the center mask, ``random'' denotes the random mask. These attention layers were implemented within our refinement framework, while using the same content generator. }
    \label{tab:attn_aware_ana}
\end{table}

\vspace{-0.2cm}\paragraph{AAL \emph{vs.} Others Context Attention Modules:} An evaluation of our proposed \emph{AAL} is shown in Table \ref{tab:attn_aware_ana}. For this quantitative experiment we used the same content generator (our TFill-\emph{Coarse}), but different attention modules in the refinement network. As can be seen, even using the same content, the proposed AAL reduces LPIPS and FID scores by averaging relative $6.0\%$ and $2.8\%$, over the existing works \cite{zhang2019self,yu2018generative,Zheng_2019_CVPR}. This is likely due to our AAL selects features based on the largest attention scores, using weights \emph{dynamically mapped} during inference, instead of depending on \emph{fixed} weights to copy features as in PIC \cite{Zheng_2019_CVPR}.

The qualitative comparison is visualized in Fig.\ \ref{fig:attn-comp}. CA \cite{yu2018generative}, PIC \cite{Zheng_2019_CVPR}, and CRFill \cite{zeng2021generative} used different context attention in image completion. Here, we directly use their publicly models for visualization. As can be seen in the Fig.\ \ref{fig:attn-comp}, these state-of-the-art methods cannot handle large holes. While TFill-\emph{SA} used the good but lower-resolution ($256\times256$) coarse content from TFill-\emph{Coarse}, the mouth exhibits artifacts with inconsistent color. Our TFill-\emph{AAL} (TFill-\emph{Refined}) shows no such artifacts.

\section{Conclusion and Limitation}

Through our analyses and experiments, we demonstrate that correctly perceiving and propagating the visible information is significantly important for masked image completion. We experimentally demonstrate the transformer-based architecture has exciting potential for content generation, due to its capacity for effectively modeling \emph{soft}-connections between distant image content. However, unlike recent vision transformer models that either use shallow projections or large receptive fields for token representation, our \emph{restrictive CNN projection} provides the necessary separation between explicit \emph{global} attention modeling and implicit \emph{local} patch correlation that leads to substantial improvement in results. We also introduced a novel attention-aware layer that adaptively balances the attention for visible and masked regions, further improving the completed image quality.

\vspace{-0.2cm}\paragraph{Limitations:} Although our TFill model outperformed existing state-of-the-art methods on various images that were degraded by random irregular masks, the model is still not able to reason about high-level semantic knowledge. For instance, while our TFill model provided better plausible results in the third row of Fig.\ \ref{fig:results_free-comp}, it directly redesigned windows based on the visible windows, without understanding the physical world, that \emph{a door is necessary for a house}. Therefore, a full understanding and imagination of semantic content in an image still needs to be further explored.

%%%%%%%%% REFERENCES
{\small
\bibliographystyle{ieee_fullname}
\bibliography{egbib}
}

\appendix\onecolumn
\renewcommand{\theequation}{\thesection.\arabic{equation}}
\setcounter{equation}{0}
\renewcommand{\thefigure}{\thesection.\arabic{figure}}
\setcounter{figure}{0}
\renewcommand{\thetable}{\thesection.\arabic{table}}
\setcounter{table}{0}
\newpage

\begin{center}
\textbf{\Large Bridging Global Context Interactions for High-Fidelity Image Completion} \\[5pt]

\textbf{\Large Supplementary Material} \\

\end{center}

The supplementary material is organized as follows: in Section~\ref{app:sec:resuls}, we first present additional visual results, including results of TFill-\emph{Coarse} on Face datasets (CelebA-HQ \cite{liu2015faceattributes,karras2018progressive}, FFHQ \cite{karras2019style}), ImageNet \cite{russakovsky2015imagenet} and Places2 datasets \cite{zhang2018perceptual}, qualitative comparison to the state-of-the-art models on various datasets, and some examples for free-form editing on high-resolution images. Next, extending the quantitative comparisons of Tables \ref{tab:conv_vs_transform} in the main paper, Section \ref{app:sec:ana} presents additional evaluation results under the traditional pixel-level and patch-level image quality metrics. Finally, we discuss more technological details of the proposed TFill model in Section \ref{app:sec:experiment}.

\section{Additional Examples}\label{app:sec:resuls}

\subsection{Additional Results for TFill-\emph{Coarse}}\label{app:sec:results_coarse}

\begin{figure}[htb!]
    \centering
    \includegraphics[width=\linewidth]{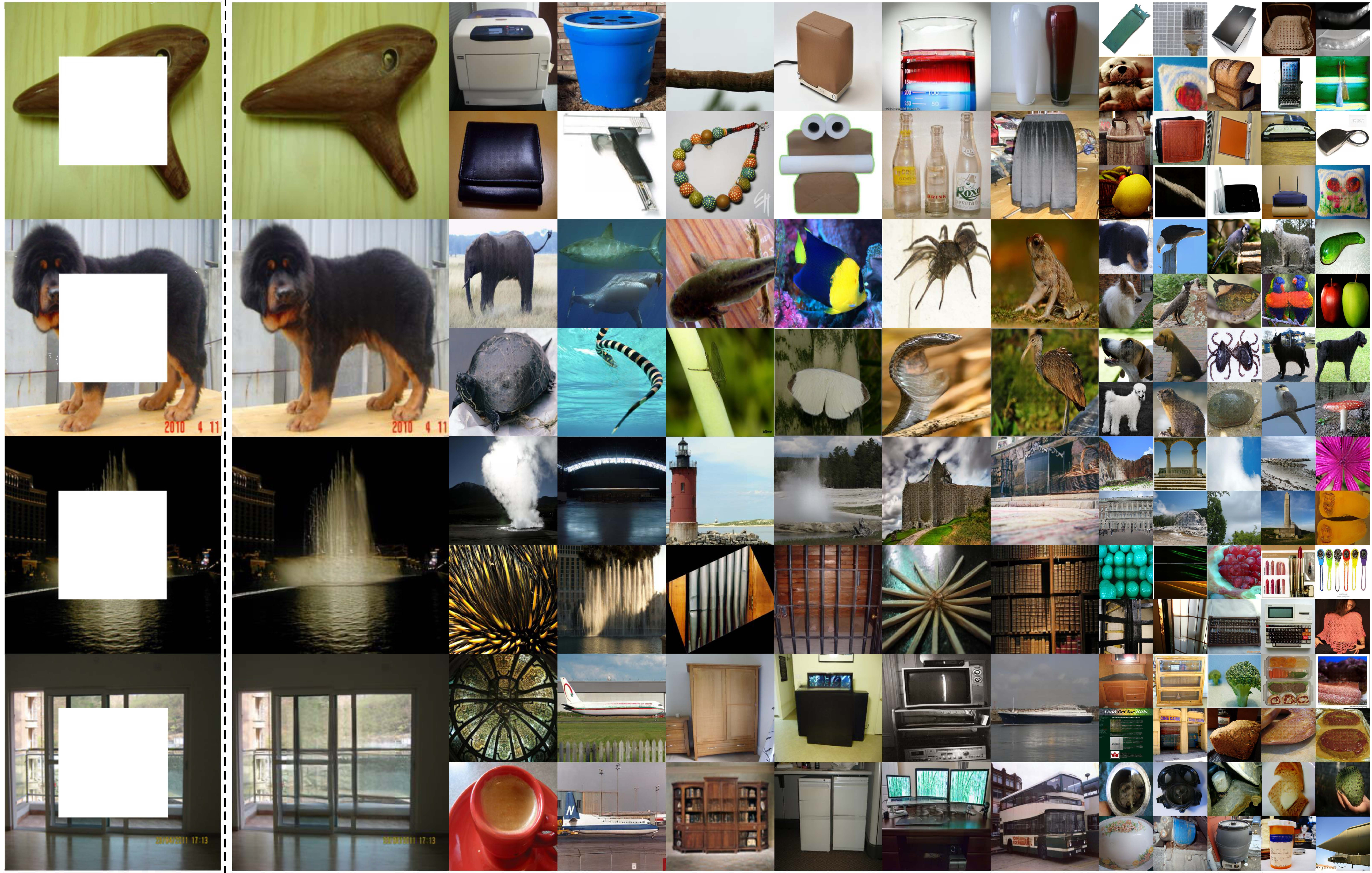}
    \caption{\textbf{Example completion results of our method (config $\mathbb{E}$) on ImageNet datasets \cite{russakovsky2015imagenet}.} All images come from the corresponding testing set that were \textbf{degraded by center masks}. Here, we show results for various categories, such as commodity, animal, plant, natural scene, building, food, furniture and so on. The center masked example inputs are shown on the left. Our model is able to complete both object shape and background scene via a transformer-based architecture to correctly bridge the visible tokens. }
    \label{fig:sup-center-imgnet-examples}
\end{figure}

\begin{figure}[tb!]
    \centering
    \includegraphics[width=\linewidth]{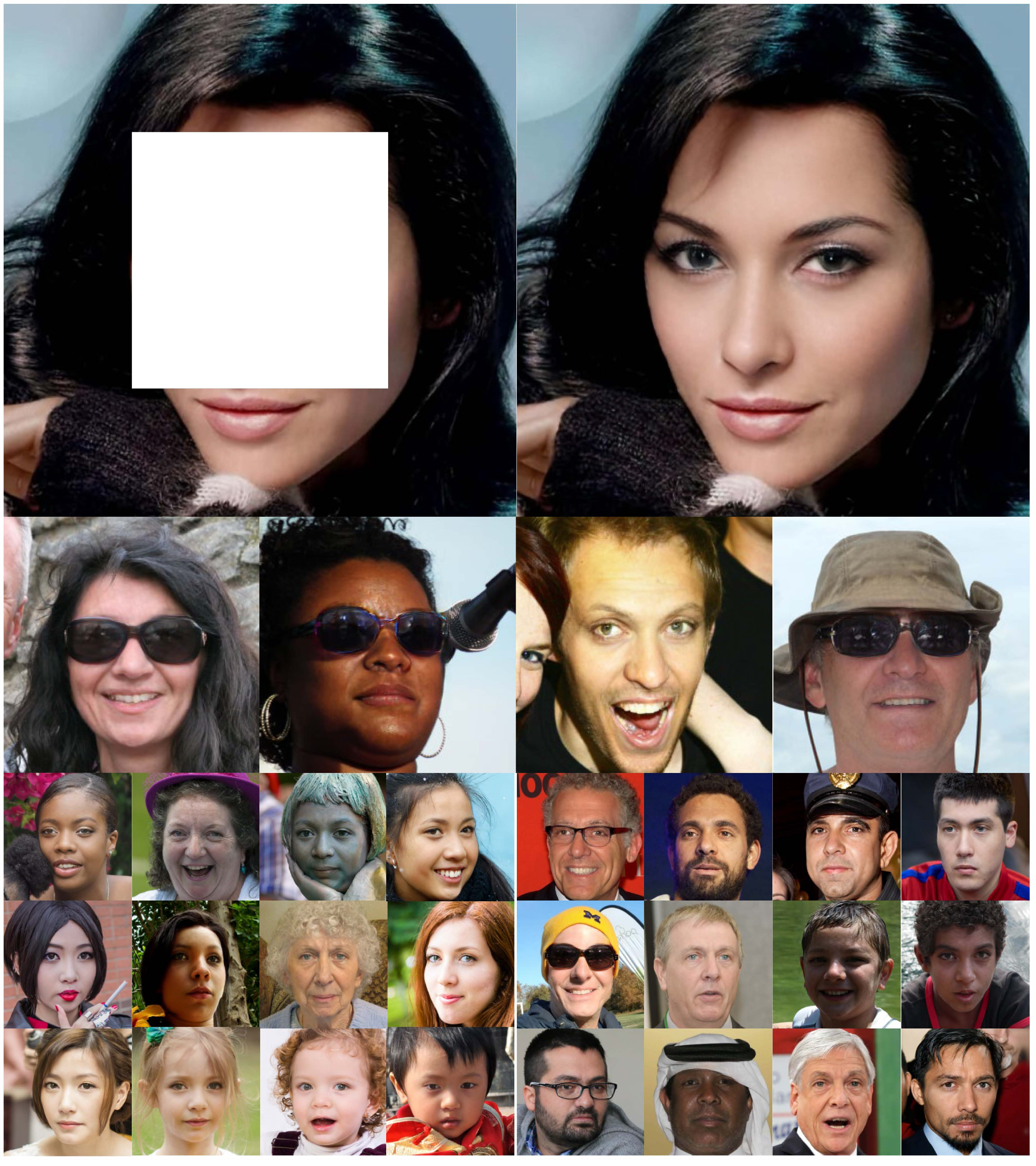}
    \caption{\textbf{Example completion results of our method (config $\mathbb{E}$) on face datasets.} Here, a center mask was used for all input images. One center masked example input is shown top-left. As can be seen, the completed images are on average of high quality. Even for some challenging cases, such as when eyeglasses are center masked, our TFill can correctly repair the face with eyeglasses. Furthermore, it generally works well for varied skin tones, poses, expressions, ages, and illumination.}
    \label{fig:sup-center-face-examples}
\end{figure}

\begin{figure}[htb!]
    \centering
    \includegraphics[width=\linewidth]{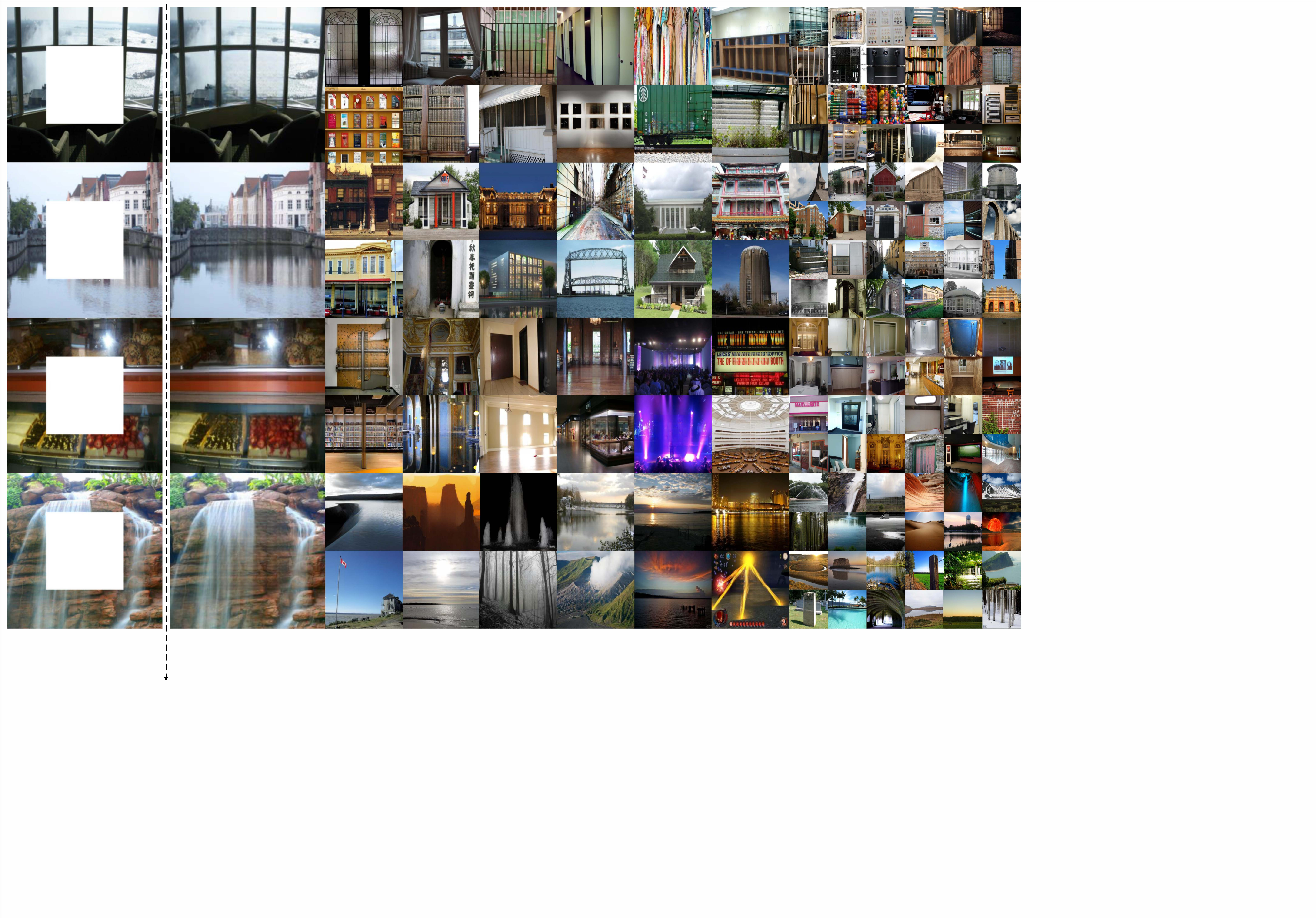}
    \caption{\textbf{Example completion results of our method (config $\mathbb{E}$) on Places2 datasets \cite{zhang2018perceptual}.} Here, we show results for varied scene categories. The center masked example inputs are shown on the left. Our model is able to complete both object shape and background scene via a transformer-based architecture to correctly bridge the visible tokens. }
    \label{fig:sup-center-places2-examples}
\end{figure}

In Figs.~\ref{fig:sup-center-imgnet-examples}, \ref{fig:sup-center-face-examples} and \ref{fig:sup-center-places2-examples}, we show more examples on ImageNet \cite{russakovsky2015imagenet}, Face datasets (CelebA-HQ \cite{liu2015faceattributes,karras2018progressive}, FFHQ \cite{karras2019style}), and Places2 \cite{zhang2018perceptual} dataset images that were degraded by large center masks. 

Here, all examples shown are chosen from the corresponding testing set. In Fig.~\ref{fig:sup-center-imgnet-examples}, we show examples for object completion, such as the various items and animals on the top half. Fig. \ref{fig:sup-center-face-examples} shows visual results of TFill-\emph{Coarse} on face datasets. In Fig.~\ref{fig:sup-center-places2-examples}, we display the completed images for various natural scenes. These examples are good evidence that our TFill model is suitable for both \emph{foreground} object completion and \emph{background} scene completion, where it can synthesize semantically consistent content with visually realistic appearance based on the presented visible pixels.

\subsection{Additional Comparisons}\label{app:sec:comparisons}

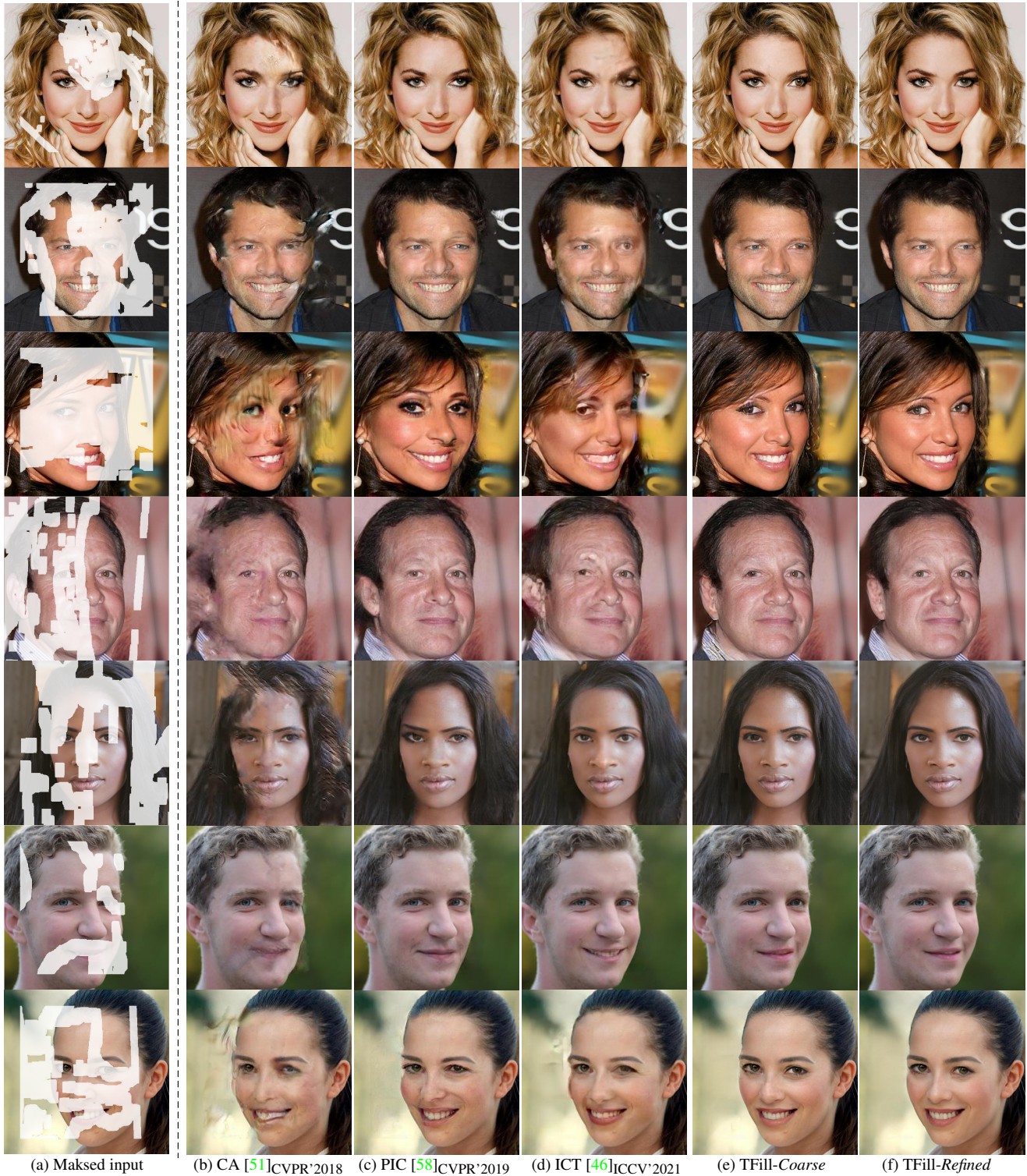
\begin{figure}[htb!]
    \centering
    \includegraphics[width=\linewidth]{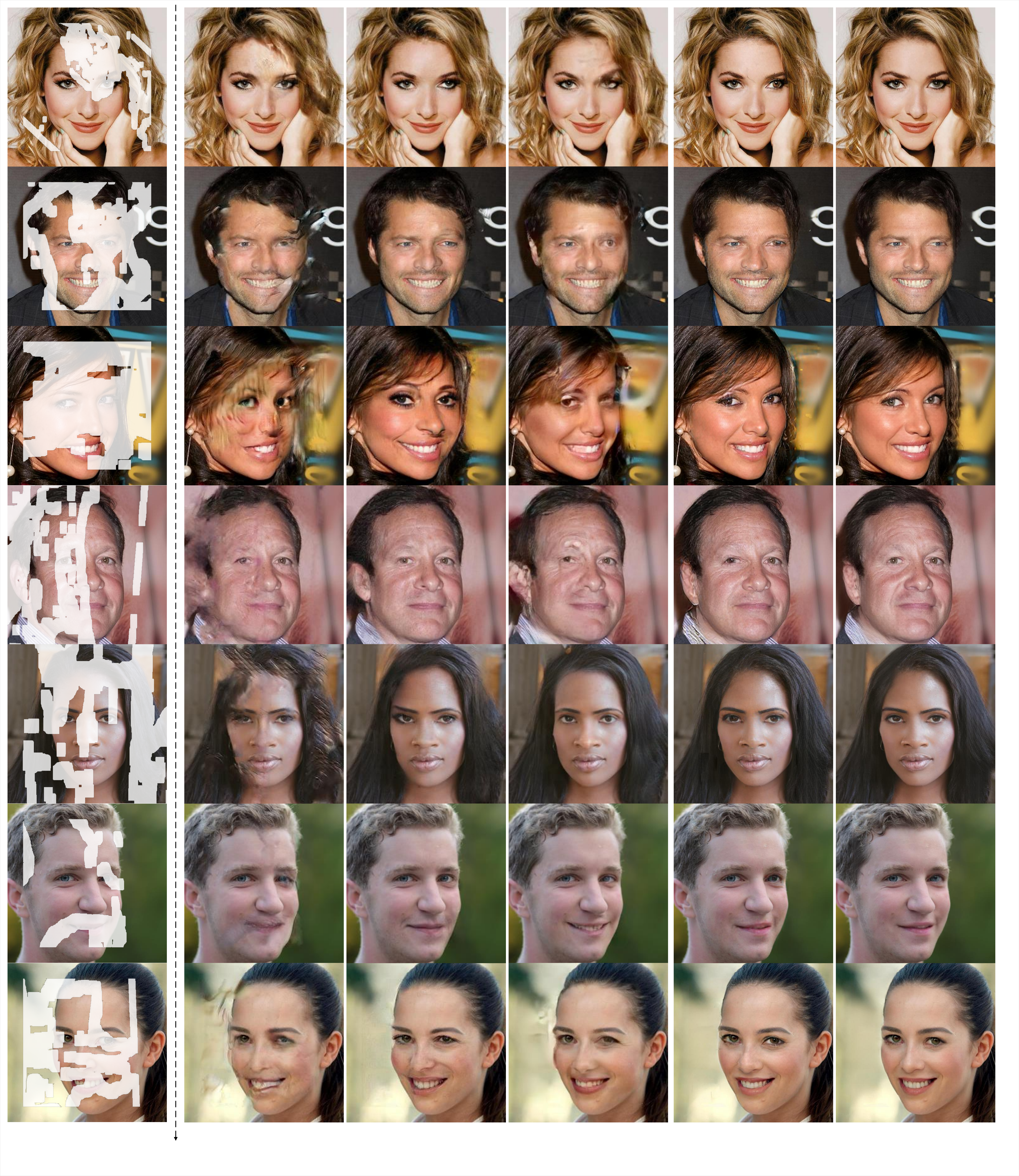}
    \begin{picture}(0,0)
    \put(-233,6){\footnotesize (a) Maksed input}
    \put(-154,6){\footnotesize (b) CA~\cite{yu2018generative}$_{\text{\scriptsize{CVPR'2018}}}$}
    \put(-74,6){\footnotesize (c) PIC~\cite{Zheng_2019_CVPR}$_{\text{\scriptsize{CVPR'2019}}}$}
    \put(10,6){\footnotesize (d) ICT~\cite{Wan_2021_ICCV}$_{\text{\scriptsize{ICCV'2021}}}$}
    \put(100,6){\footnotesize (e) TFill-\emph{Coarse}}
    \put(180,6){\footnotesize (f) TFill-\emph{Refined}}
    \end{picture}
    \vspace{-0.2cm}
    \caption{\textbf{Additional results on CelebA-HQ \cite{liu2015faceattributes,karras2018progressive} and FFHQ \cite{karras2019style} testing set among CA~\cite{yu2018generative}, PIC~\cite{Zheng_2019_CVPR}, ICT~\cite{Wan_2021_ICCV} and Ours.} Our results are reported for $512^2$ resolution. While PIC~\cite{Zheng_2019_CVPR} works well for frontal facing faces, it may generate more uncanny faces with mismatched features at larger angles, \eg the examples in third and last row. In contrast, our model generated consistent facial features with photorealistic appearance for various faces angles. As ICT \cite{Wan_2021_ICCV} does not use context attention in the second stage to copy information from visible pixels, most of eyes on the completed images are inconsistent. Zoom in to see the details.}
    \label{fig:sup-free-form-face}
\end{figure}

\begin{figure}[htb!]
    \centering
    \includegraphics[width=\linewidth]{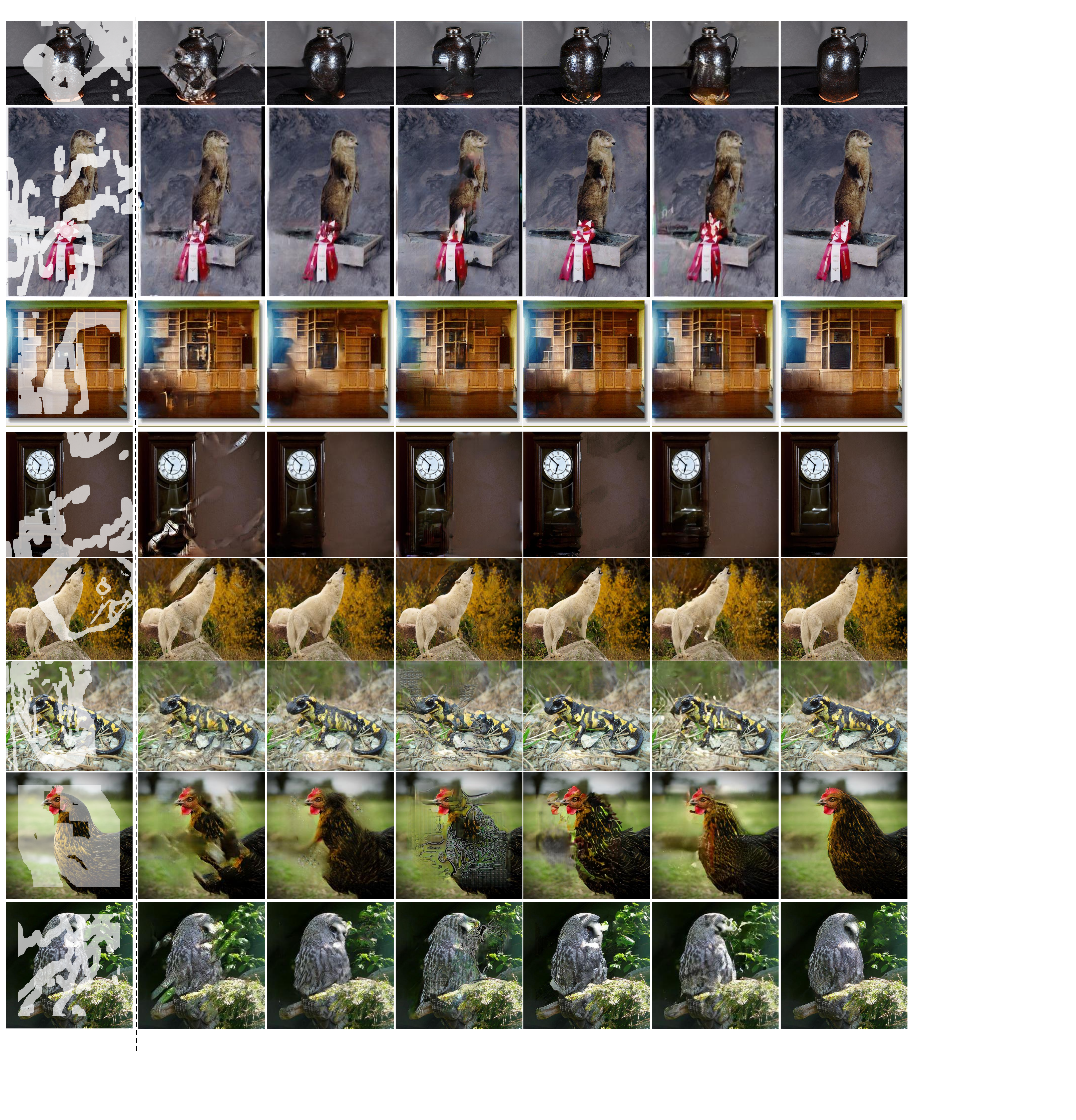}
    \begin{picture}(0,0)
    \put(-238,4){\footnotesize (a) Masked input}
    \put(-172,4){\footnotesize (b) CA~\cite{yu2018generative}$_{\text{\scriptsize{CVPR'18}}}$}
    \put(-104, 4){\footnotesize (c) PIC~\cite{Zheng_2019_CVPR}$_{\text{\scriptsize{CVPR'19}}}$}
    \put(-36, 4){\footnotesize (d) HiFill~\cite{yi2020contextual}$_{\text{\scriptsize{CVPR'20}}}$}
    \put(36, 4){\footnotesize (e) CRFill~\cite{zeng2021generative}$_{\text{\scriptsize{CVPR'21}}}$}
    \put(112, 4){\footnotesize (f) ICT~\cite{Wan_2021_ICCV}$_{\text{\scriptsize{ICCV'21}}}$}
    \put(200,4){\footnotesize (g) TFill}
    \end{picture}
    \caption{\textbf{Additional results on ImageNet \cite{russakovsky2015imagenet} testing set among CA~\cite{yu2018generative}, PIC~\cite{Zheng_2019_CVPR}, HiFill~\cite{yi2020contextual}, CRFill~\cite{zeng2021generative}, ICT~\cite{Wan_2021_ICCV} and Ours.} Our results are evaluated in higher resolution, with the short side at $512$ pixels and the long side at multiples of $2^5$, \eg 640. Our TFill model generated better visual results even under very challenging situations, \eg the heavily masked chicken in the second last row. }
    \label{fig:sup-free-form-imagenet}
\end{figure}
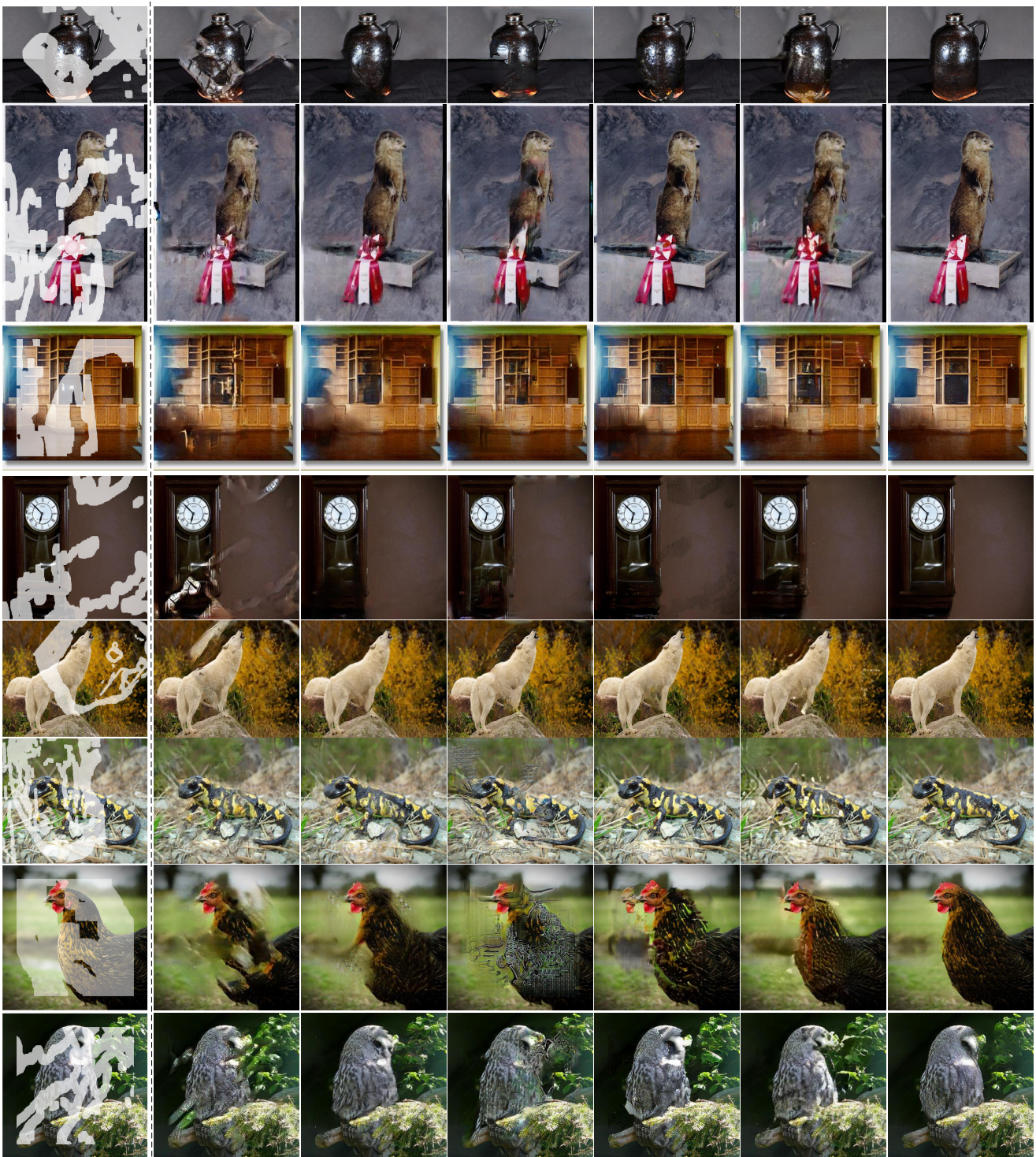

\begin{figure}[htb!]
    \centering
    \includegraphics[width=\linewidth]{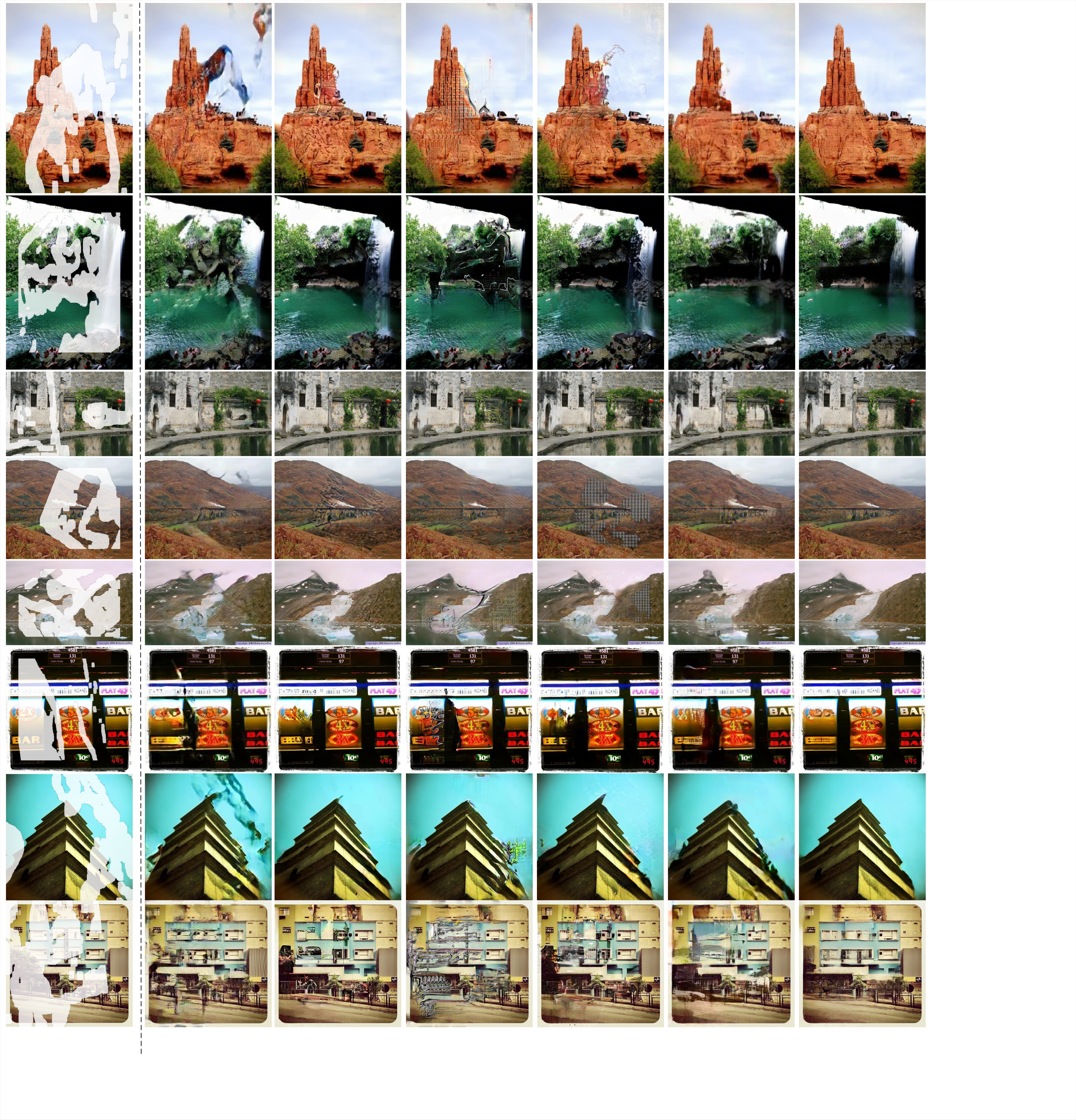}
    \begin{picture}(0,0)
    \put(-238,4){\footnotesize (a) Masked input}
    \put(-172,4){\footnotesize (b) CA~\cite{yu2018generative}$_{\text{\scriptsize{CVPR'18}}}$}
    \put(-104, 4){\footnotesize (c) PIC~\cite{Zheng_2019_CVPR}$_{\text{\scriptsize{CVPR'19}}}$}
    \put(-36, 4){\footnotesize (d) HiFill~\cite{yi2020contextual}$_{\text{\scriptsize{CVPR'20}}}$}
    \put(36, 4){\footnotesize (e) CRFill~\cite{zeng2021generative}$_{\text{\scriptsize{CVPR'21}}}$}
    \put(112, 4){\footnotesize (f) ICT~\cite{Wan_2021_ICCV}$_{\text{\scriptsize{ICCV'21}}}$}
    \put(200,4){\footnotesize (g) TFill}
    \end{picture}
    \caption{\textbf{Additional results on Places2 \cite{zhang2018perceptual} testing set among CA~\cite{yu2018generative}, PIC~\cite{Zheng_2019_CVPR}, HiFill~\cite{yi2020contextual}, CRFill~\cite{zeng2021generative}, ICT~\cite{Wan_2021_ICCV} and Ours}. Our results are evaluated in higher resolution, with the short side at $512$ pixels and the long side at multiples of $2^5$, \eg 640. }
    \label{fig:sup-free-form-place}
\end{figure}

In Figs.~\ref{fig:sup-free-form-face}, \ref{fig:sup-free-form-imagenet} and \ref{fig:sup-free-form-place}, we show additional comparison results on various datasets with free-form masks provided in PConv~\cite{Liu_2018_ECCV}. This is an extension of Figs.~\ref{fig:results_free-comp} and \ref{fig:attn-comp} in the main paper. Here, our results on CelebA-HQ \cite{liu2015faceattributes,karras2018progressive} and FFHQ \cite{karras2019style} testing set are reported for $512\times512$ resolution. On the other hand, the results on ImageNet \cite{russakovsky2015imagenet} and Places2 \cite{zhang2018perceptual} are reported for higher resolution images that were resized such that the short side is 512 pixels, with the long side in multiples of $2^5=32$, \eg $640=32 \times 20$. The size variability is possible due to our fully convolutional encoder-decoder network structures. The 32-base scale is required because our refinement network downsamples the images 5 times with step 2. 

As can be seen from these results, our TFill model filled appropriate semantic content with visually realistic appearance into the various masks. For instance, in the third row of Fig.~\ref{fig:sup-free-form-face}, even with an extensive mask on an obliquely angled face, it was able to generate high-quality results. It achieved good results even under challenging conditions for various objects (Fig.~\ref{fig:sup-free-form-imagenet}) and scenes (Fig.~\ref{fig:sup-free-form-place}). 

\subsection{Free-Form Editing on High-Resolution Images}\label{app:sec:removing}

\begin{figure}[htb!]
    \centering
    \includegraphics[width=\linewidth]{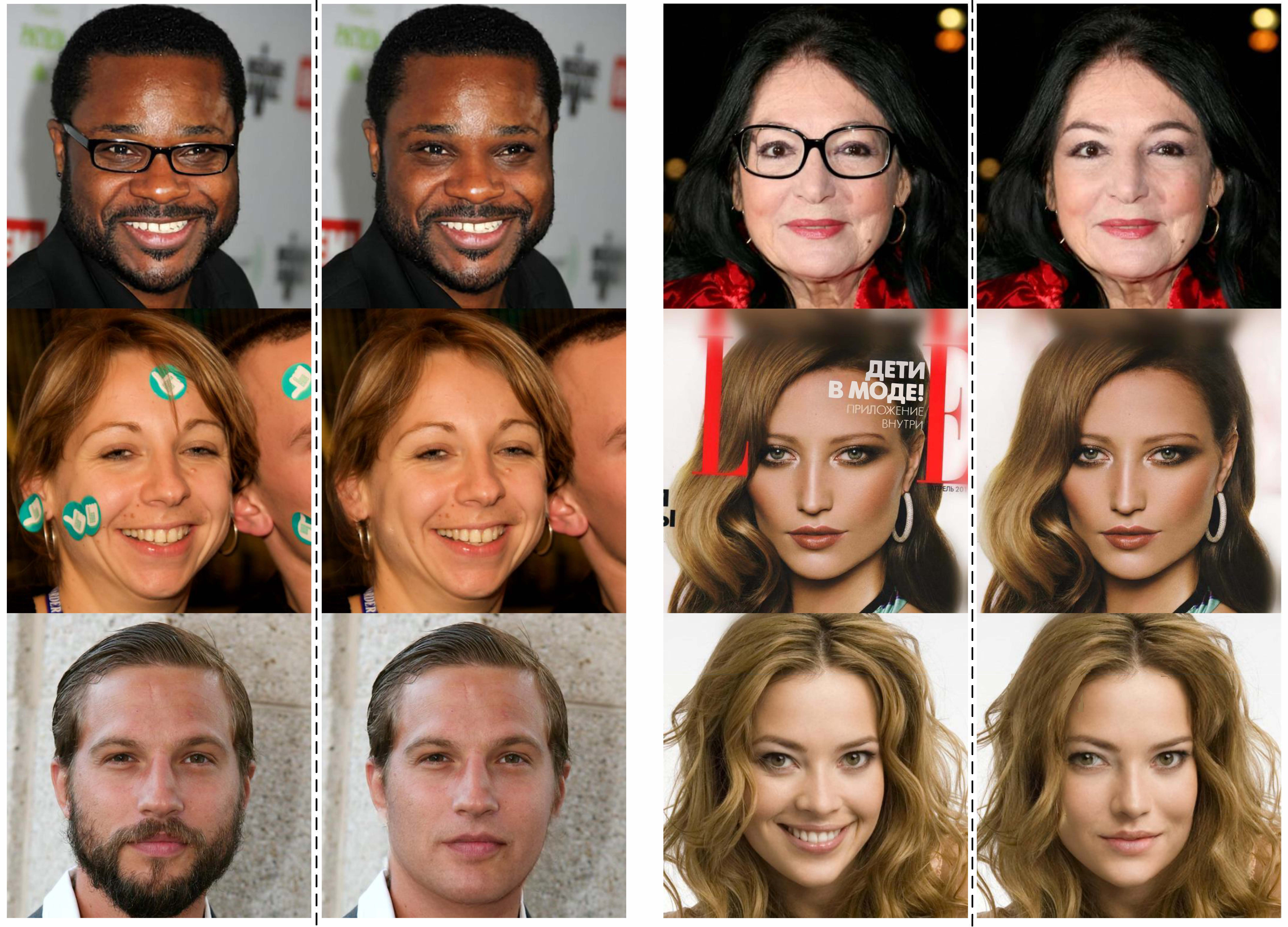}
    \begin{picture}(0,0)
    \put(-204,6){\footnotesize (a) Original}
    \put(-94,6){\footnotesize (b) Edited Output}
    \put(52,6){\footnotesize (a) Original}
    \put(164,6){\footnotesize (b) Edited Output}
    \end{picture}
    \vspace{-0.2cm}
    \caption{\textbf{Additional results on CelebA-HQ \cite{liu2015faceattributes,karras2018progressive} and FFHQ \cite{karras2019style} testing set for free-form mask editing.} All results are reported at $512^2$ resolution. Our model works well for traditional object removal, such as removing eyeglasses and watermarks. Furthermore, we provide  examples of more substantial modifications, \eg facial hair removal, and expression modification in the last row.}
    \label{fig:sup-free-edit-face}
\end{figure}

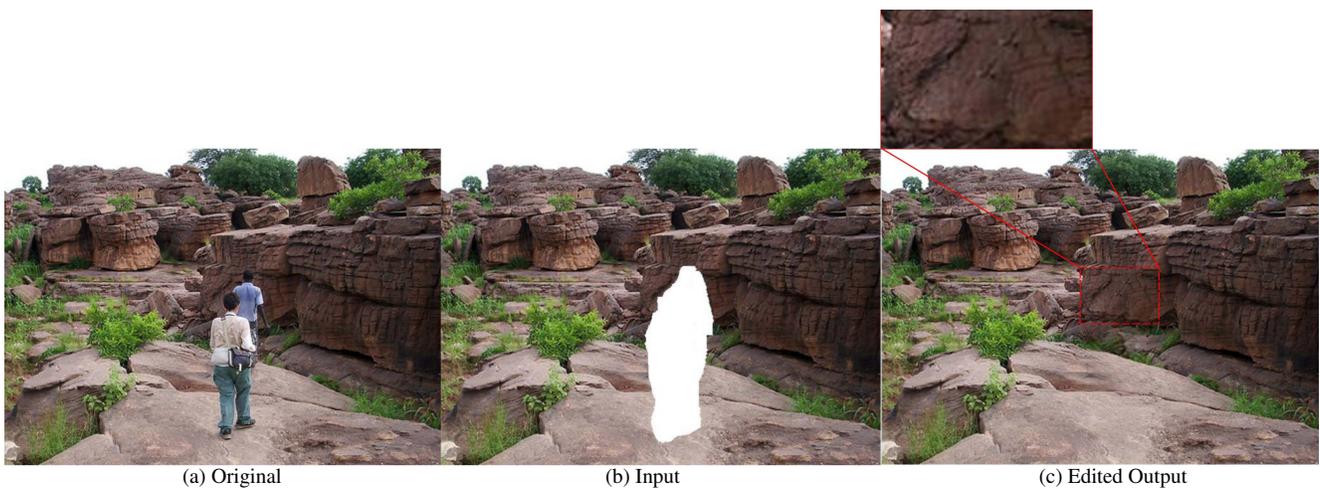
\begin{figure}[htb!]
    \centering
    \includegraphics[width=\linewidth,height=0.27\textheight]{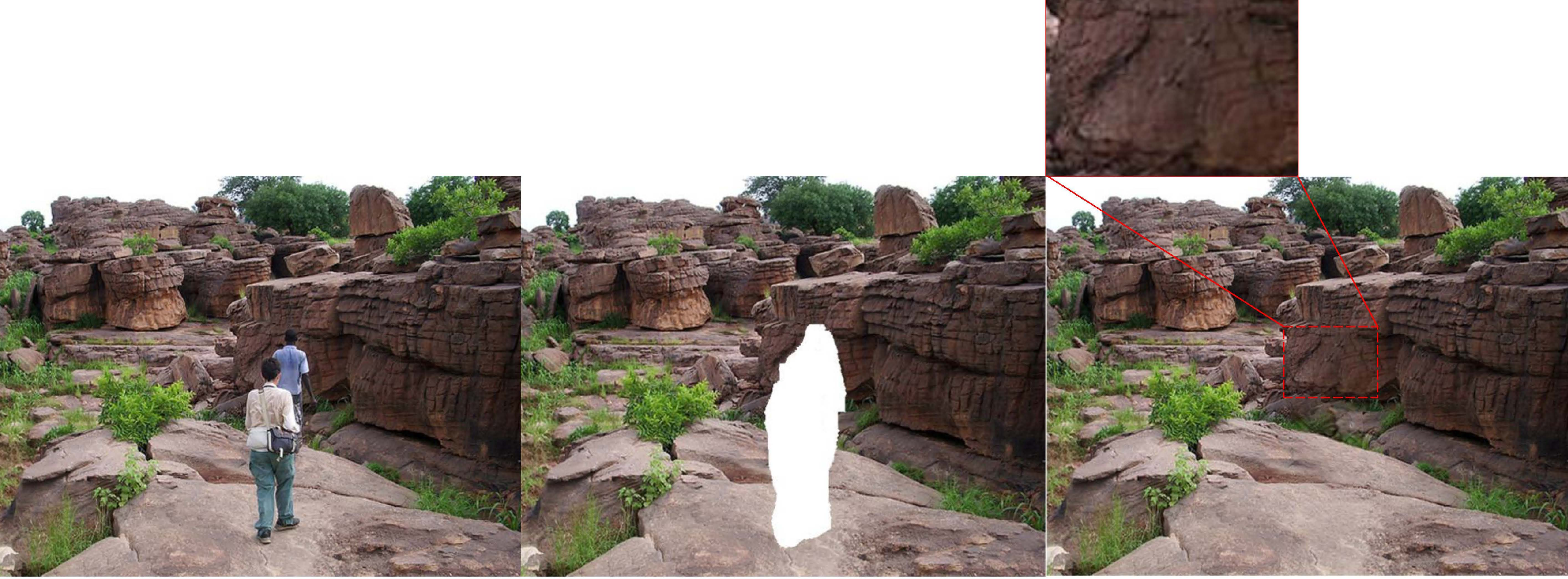}
    \begin{picture}(0,0)
    \put(-180,6){\footnotesize (a) Original}
    \put(-20,6){\footnotesize (b) Input}
    \put(144,6){\footnotesize (c) Edited Output}
    \end{picture}
    \vspace{-0.2cm}
    \caption{\textbf{Additional free-form editing results on ImageNet \cite{russakovsky2015imagenet}.} The original image size in ImageNet is $500\times375$. Here, we resized slightly to $512\times384$ for image completion. We highlight the generated content, which has consistent textures to those in the visible regions.}
    \label{fig:sup-free-edit-imagenet0}
\end{figure}

\begin{figure}[tb!]
    \centering
    \includegraphics[width=\linewidth]{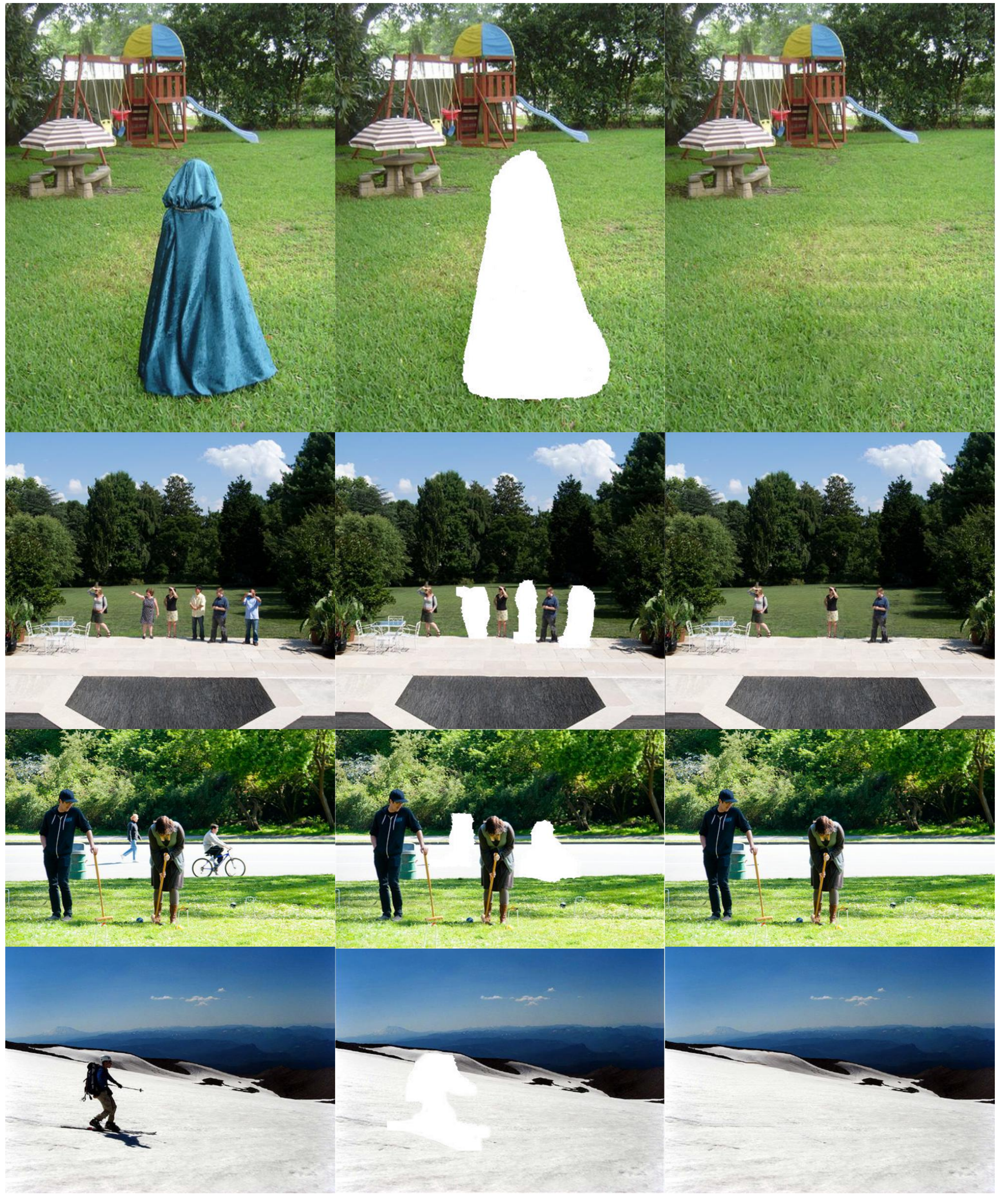}
    \begin{picture}(0,0)
    \put(-180,8){\footnotesize (a) Original}
    \put(-20,8){\footnotesize (b) Input}
    \put(144,8){\footnotesize (c) Edited Output}
    \end{picture}
    \vspace{-0.2cm}
    \caption{\textbf{Additional results on ImageNet \cite{russakovsky2015imagenet} testing set for free-form editing.} Here, we enforce the input image size to be multiples of $32$, \eg $960\times640$ and provide the high-resolution results on the corresponding image size. Zoom in to see the completed details.}
    \label{fig:sup-free-edit-imagenet}
\end{figure}

\begin{figure}[tb!]
    \centering
    \includegraphics[width=\linewidth]{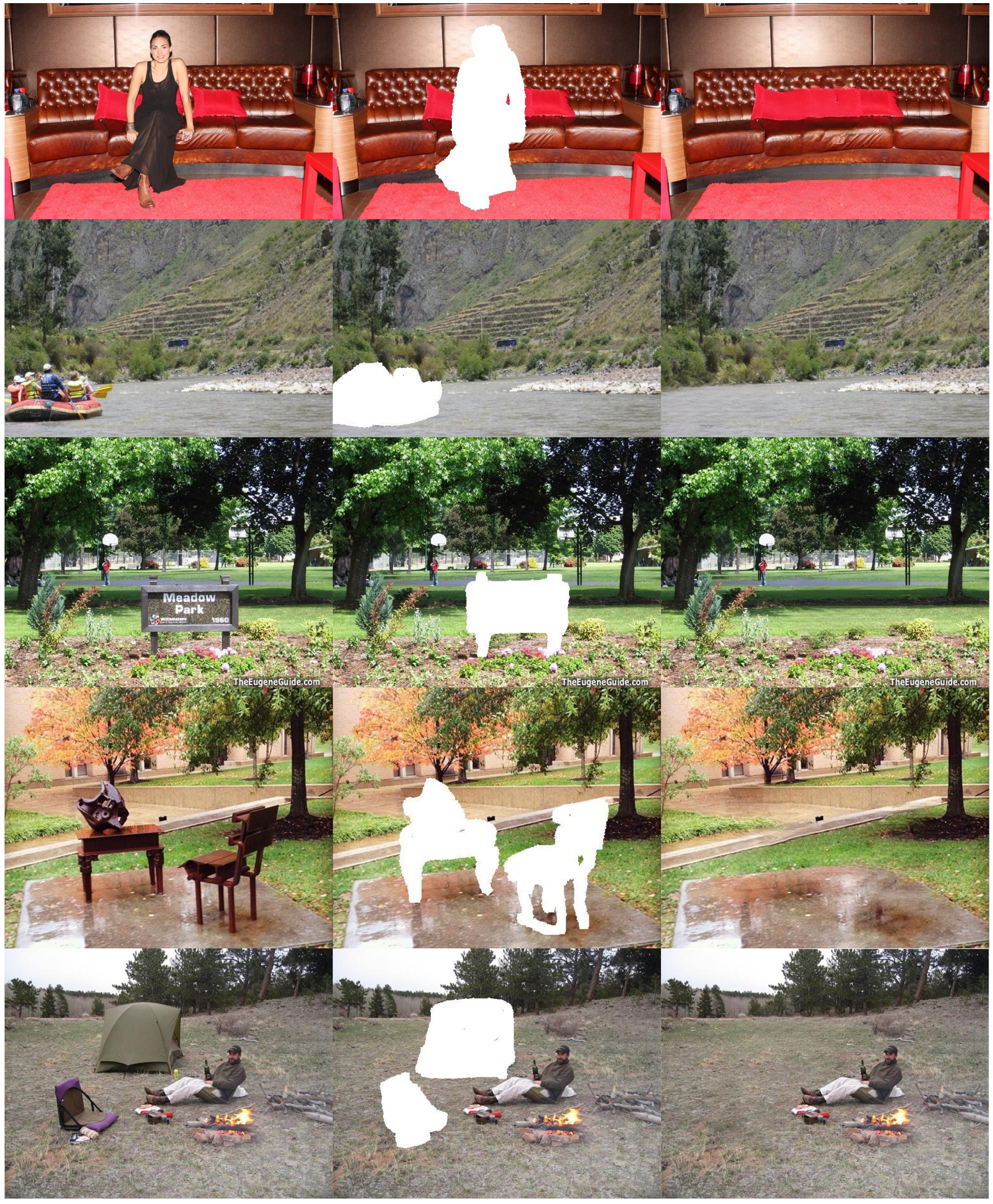}
    \begin{picture}(0,0)
    \put(-180,6){\footnotesize (a) Original}
    \put(-20,6){\footnotesize (b) Input}
    \put(144,6){\footnotesize (c) Edited Output}
    \end{picture}
    \vspace{-0.2cm}
    \caption{\textbf{Additional results on Places2 \cite{zhang2018perceptual} testing set for free-form editing.} Here, we enforce the input image size to be multiples of $32$, \eg $960\times640$ and provide the high-resolution results on the corresponding image size. Zoom in to see the completed details.}
    \label{fig:sup-free-edit-places}
\end{figure}

In Figs.\ \ref{fig:sup-free-edit-face}, \ref{fig:sup-free-edit-imagenet0}, \ref{fig:sup-free-edit-imagenet} and \ref{fig:sup-free-edit-places}, we show qualitative results for free-form image masking on various higher resolution datasets. 

In Fig.~\ref{fig:sup-free-edit-face}, we show some examples for face editing at $512^2$ resolution. For conventional object removal, \eg watermark removal, our TFill addresses them easily. Furthermore, our TFill can handle more extensive face editing, such as removing substantial facial hair and changing mouth expressions.

In Figs.~\ref{fig:sup-free-edit-imagenet0}, \ref{fig:sup-free-edit-imagenet} and \ref{fig:sup-free-edit-places}, we show some examples of editing images of natural / outdoor scenes, with object removal being the main task, as it is the main practice for image inpainting. Here, we enforce the input image size to be multiples of $32$, \eg $960\times640$ and provide the high-resolution results on the corresponding image size. As we can see, our TFill-\emph{Refined} model is able to handle high-resolution images for object removal in traditional image inpainting task. 

\clearpage
\newpage

\section{Additional Quantitative Results}\label{app:sec:ana}

We further report quantitative results using traditional pixel-level and patch-level image quality evaluation metrics. 

\begin{table*}[htb!]
    \centering
    \renewcommand{\arraystretch}{1.1}
    \setlength\tabcolsep{4pt}
    \begin{tabular}{@{}llccccccc@{}}
         \hlineB{3.5}
         & \multirow{2}{*}{\textbf{Method}} & \multicolumn{3}{c}{\textbf{CelebA-HQ}}&& \multicolumn{3}{c}{\textbf{FFHQ}}\\
		\cline{3-5}\cline{7-9}
		& & $\ell_1 \text{loss}\downarrow$& SSIM$\uparrow$ & PSNR$\uparrow$ && $\ell_1 \text{loss}\downarrow$ & SSIM$\uparrow$ & PSNR$\uparrow$ \\
		\hlineB{2}
		& CA~\cite{yu2018generative} & 0.0310 & 0.8201 & 23.5667 && 0.0337 & 0.8099 & 22.7745 \\
		& PIC~\cite{Zheng_2019_CVPR} & 0.0209 & 0.8668 & 24.6860 & & 0.0241 & 0.8547 & 24.3430\\
		& MEDFE~\cite{Liu2019MEDFE} & 0.0208 & 0.8691 & 24.4733 & & -& - & - \\
		\cdashline{1-7}
		$\mathbb{A}$ & Traditional \emph{Conv} & 0.0199 & 0.8693 & 24.5800 && 0.0241 & 0.8559 & 24.2271 \\
		$\mathbb{B}$ & + Attention in G & 0.0196 & 0.8717 & 24.6512 && 0.0236 & 0.8607 & 24.4384 \\
		$\mathbb{C}$ & + Restrictive \emph{Conv} & 0.0191 & 0.8738 & 24.8067 && 0.0220 & 0.8681 & 24.9280 \\
		$\mathbb{D}$ & + Transformer & 0.0189 & 0.8749 & 24.9467 && 0.0197 & 0.8751 & 25.1002\\
		$\mathbb{E}$ & + Masked Attention & 0.0183 & 0.8802 & 25.2510 && 0.0188 & 0.8765 & 25.1204 \\
		$\mathbb{F}$ & + Refine Network & \textbf{0.0180} & \textbf{0.8821} & \textbf{25.4220} && \textbf{0.0184} & \textbf{0.8778}& \textbf{25.2061} \\
        \hlineB{2}
    \end{tabular}
    \caption{Quantitative results for traditional pixel-level and patch-level metrics on center masked images. }
    \label{tab:sup_conv_vs_transform}
\end{table*}

Table \ref{tab:sup_conv_vs_transform} provides a comparison of our results to state-of-the-art CNN-based models, as well as various alternative configurations for our design, on the center masked face testing set. This is an extension of Table \ref{tab:conv_vs_transform} in the main paper. All images were normalized to the range [0,1] for quantitative evaluation. While there is no necessity to strongly encourage the completed images to be the same as the original ground-truth images, our TFill model nonetheless achieved better performance on these metrics too, including $\ell_1$ loss, structure similarity index (SSIM) and peak signal-to-noise ration (PSNR), suggesting that our TFill model is more capable of generating closer content to the original unmasked images. 

\section{Experiment Details}
\label{app:sec:experiment}

Here we first present the novel layers and loss functions used to train our model, followed by the training details. 

\subsection{Multihead \emph{Weighted} Self-Attention}
\label{app:sec:msa}

Our transformer encoder is built on the standard \textbf{qkv} \texttt{self-attention} (SA) \cite{Vaswani_NIPS2017_attention} with a learned position embedding in each layer. Given an input sequence $\textbf{z}\in\mathbb{R}^{N\times C}$, we first calculate the pairwise similarity $\textbf{A}$ between each two elements as follows:
\begin{align}
    [\textbf{q}, \textbf{k}, \textbf{v}] & = \textbf{W}_{qkv}\textbf{z} \\
    \textbf{A} & = \texttt{softmax}(\textbf{q}\textbf{k}^{\top}/\sqrt{C_h})
\end{align}
where $\textbf{W}_{qkv}\in\mathbb{R}^{C\times3C_{h}}$ is the learned parameter to refine the features \textbf{z} for the query $\textbf{q}$, the key $\textbf{k}$ and the value $\textbf{v}$. $\textbf{A}\in\mathbb{R}^{N\times N}$ is the dot similarity of N tokens, which is scaled by the square root of feature dimension $C_h$. Then, we compute a weighted sum over all values $\textbf{v}$ via:
\begin{equation}
    \text{SA}(\textbf{z}) = \textbf{A}\textbf{v}
\end{equation}
where the value $z$ in the sequence is connected through their learned similarity $A$, rather than purely depending on a fixed learned weight $w$. 

The multihead \texttt{self-attention} (MSA) is an extention of SA, in which $H$ heads are run in parallel to get multiple attention scores and the corresponding projected results. Then we get the following function:
\begin{equation}
    \text{MSA}(\textbf{z})=[\text{SA}_1(\textbf{z});\text{SA}_2(\textbf{z});\dots;\text{SA}_h(\textbf{z})]
\end{equation}

To encourage the model to \emph{bias} to the important visible values, we further modify the MSA with a \emph{masked} self-attention layer, in which a masked weight is applied to scale the attention score $\textbf{A}$. Given a feature $\textbf{x}$ and the corresponding mask $\textbf{m}$ (1 denotes visible pixel and 0 is masked pixel). The original partial convolution operation is operated as:
\begin{align}
    x^\prime & = 
    \begin{cases}
    \textbf{W}_p(\textbf{x}_p\bigodot\textbf{m}_p)\frac{1}{\sum(\textbf{m}_p)} + b, & \mbox{if} \sum(\textbf{m}_p) > 0 \\
    0, & \mbox{otherwise}
    \end{cases} \\
    m^\prime & = 
    \begin{cases}
    1, & \mbox{if} \sum(\textbf{m}_p) > 0 \\
    0, & \mbox{otherwise}
    \end{cases}
\end{align}
where $\textbf{W}_p$ contain the convolution filter weights, $b$ is the corresponding bias, while $\textbf{x}_p$ and $\textbf{m}_p$ are the feature values and mask values in the current convolution window (\eg $2\times2$ in our \emph{restrictive CNN}), respectively. Here, we replace the $m^\prime$ as a float value:
\begin{equation}
    m^\prime = \frac{\sum(\textbf{m}_p)}{S}
\end{equation}
where $S$ is the size of each convolution filter, $2\times2$ used in our \emph{restrictive CNN}. To do this, each token only extracts the visible information. What's more, the final $m$ for each token denotes the percentage of valid values in each token under a small RF. Then, for each sequence $\textbf{z}\in\mathbb{R}^{N\times C}$, we obtain a corresponding masked weight $\textbf{m}\in\mathbb{R}^{N\times 1}$ by flattening the updated mask. Finally, we update the original attention score by multiplying with the repeated masked weight $\textbf{m}\in\mathbb{R}^{N\times 1}$:
\begin{equation}
    \textbf{A}_m = \textbf{A}\bigodot\textbf{m}_r
\end{equation}
where $\textbf{m}_r\in\mathbb{R}^{N\times N}$ is the extension of masked weight $\textbf{m}\in\mathbb{R}^{N\times 1}$ in the final dimension. 

\subsection{Loss Functions}\label{app:sec:loss}

Our work focuses on exploiting the \emph{token representation} in the visual transformer architecture. We do \emph{not} modify the discriminator architecture or design the loss function in any way. Both TFill-\emph{Coarse} and TFill-\emph{Refined} is trained with loss $L = L_{pixel} + L_{per} + L_{GAN}$. In particular, each loss is given as:
\begin{align}
    L_{pixel} & = ||\textbf{I}_{gt}-\textbf{I}_g||_1 \\
    L_{per} & = ||\Phi_{n}(\textbf{I}_{gt})-\Phi_{n}(\textbf{I}_g)||_1 \\
    L_{GAN} & = log(1+\exp(-D(\textbf{I}_g)))
\end{align}
where $\textbf{I}_g$ and $\textbf{I}_{gt}$ is the generated image and original ground truth image, respectively. $\Phi_{n}$ is the activation map of the $n$th selected layer in VGG. $D$ is the discriminator and here we show only the generator loss for the generative adversarial traning. 

\subsection{Training Details}\label{app:sec:training}

Our model was trained on two NVIDIA A100 GPUs in two stages: \textbf{1)} the content inference network was first trained with $256^2$ resolution with batch size of 96; \textbf{2)} the visual appearance network was then trained with $512^2$ resolution with batch size of 24. Both networks were optimized using the loss $L = L_{pixel} + L_{per} + L_{GAN}$. The design of the encoder-decoder backbone follows the architecture presented in \cite{esser2020taming}. For the discriminator, we used the architecture of StyleGANv2 \cite{karras2020analyzing}. 

\end{document}